\documentclass[journal]{IEEEtran}

\usepackage[T1]{fontenc}
\usepackage[utf8]{inputenc}
\usepackage{graphicx}
\usepackage{textcomp}
\usepackage{gensymb}
\usepackage{amssymb}
\usepackage{orcidlink}
\usepackage[numbers]{natbib}
\usepackage{subcaption}

\usepackage{colortbl}
\usepackage{tabularx}
\usepackage{array}

\usepackage{soul}
\usepackage{xcolor} 

\begin{document}

\title{A systematic review of the use of Deep Learning in Satellite Imagery for Agriculture}

\author{Brandon Victor \orcidlink{0000-0001-5832-6087}, Aiden Nibali \orcidlink{0000-0003-0302-5775}, Zhen He \orcidlink{0000-0002-1352-9910}
\thanks{Brandon Victor, Aiden Nibali and Zhen He work within the School of Computing, Engineering and Mathematical Sciences at La Trobe University, Melbourne, Victoria, Australia}}

\maketitle

\begin{abstract}
Agricultural research is essential for increasing food production to meet the needs of a rapidly growing human population. Collecting large quantities of agricultural data helps to improve decision making for better food security at various levels: from international trade and policy decisions, down to individual farmers. At the same time, deep learning has seen a wave of popularity across many different research areas and data modalities. And satellite imagery has become available in unprecedented quantities, driving much research from the wider remote sensing community. The data hungry nature of deep learning models and this huge data volume seem like a perfect match. But has deep learning been adopted for agricultural tasks using satellite images? This systematic review of 193 studies analyses the tasks that have reaped benefits from deep learning algorithms, and those that have not. It was found that while Land Use / Land Cover research has embraced deep learning algorithms, research on other agricultural tasks has not. This poor adoption appears to be due to a critical lack of labelled datasets for these other tasks. Thus, we give suggestions for collecting larger datasets. Additionally, satellite images differ from ground-based images in a number of ways, resulting in a proliferation of interesting data interpretations unique to satellite images. So, this review also introduces a taxonomy of data input shapes and how they are interpreted in order to facilitate easier communication of algorithm types and enable quantitative analysis.
\end{abstract}








\section{Introduction}

\IEEEPARstart{T}{here} are big agricultural challenges coming. The human population is expected to increase significantly in the next decades \cite{united_nations_world_2019}, which will require an estimated global yield increase of 25-70\% \cite{hunter_agriculture_2017}. At the same time, a changing climate brings additional challenges \cite{tirado_climate_2010} and the need to reduce the environmental impact of agriculture \cite{gebbers_precision_2010}.

Collecting more agricultural data is one path to improving global food production and distribution. Remote sensing is an extremely useful tool for such because it provides a non-destructive and non-intrusive way to monitor agricultural fields simultaneously at a fine level of detail and across wide areas and times. This makes it a technology that can be used for fast and targeted interventions at a farm-level \cite{griffin_worldwide_2005}, regional studies of ecological change over time \cite{roughgarden_what_1991}, county-level yield prediction for logistics \cite{fritz_comparison_2019} and international trade decisions \cite{macdonald_global_1980}. To achieve these, there are many sources of worldwide satellite imagery freely available to the public. The most popular for agricultural purposes are: Sentinel \cite{torres_gmes_2012, drusch_sentinel-2_2012}, Landsat \cite{roy_landsat8_2014} and MODIS. Each of these satellite programs store and manage enormous collections of historical worldwide imagery e.g. Sentinel added 7.34 PiB of imagery to their archive in 2021 \cite{serco_copernicus_2021}. The availability of this data is recognised as a key driver of research \cite{miller_users_2013}.

At the same time, deep learning has become the dominant approach in all generic computer vision dataset competitions \cite{deng_imagenet_2009, lin_microsoft_2014, kuznetsova_open_2020}. On those tasks, deep learning outperforms traditional feature engineering and machine learning by a wide margin. It is generally agreed that these successes are only possible because of the availability of large training datasets \cite{goodfellow_deep_2016}. Thus, given the large volumes of data available for remote sensing, a similar shift is expected in remote sensing algorithms. But has this been the case for agricultural tasks?

This systematic review of 193 studies investigates and quantifies the use of deep learning on satellite images across agricultural tasks. It was found that there were very few examples of modern deep learning methods from before 2020, after which they have become increasingly popular, with an explosion of research in just the last few years. However, this increase in popularity has been mostly for Land Use / Land Cover (LULC) tasks, and to a lesser extent in yield prediction. Research into other agricultural tasks have not seen the same level of adoption. Consequently, there was a wide variety of approaches taken for LULC tasks, but the most common approach for the other tasks is still a pixel-based Random Forest (RF) or Multilayer Perceptron (MLP). Nevertheless, it was found that where spatial Convolutional Neural Networks (CNNs) were used, they consistently outperformed traditional machine learning methods (ML) across all tasks. However, Long Short-Term Memory models (LSTMs) did not consistently outperform traditional ML method for yield prediction. There were few papers that included attention-based models (both ViT \cite{vaswani_attention_2017} and custom architectures \cite{garnot_satellite_2020}), but there was no consistent improvement observed over other modern deep learning techniques in the reviewed studies.

Compared to ground-level images, satellite images have lower spatial resolutions, higher spectral resolution and are typically processed to obtain reflectances (a physical measurement). Labels for remote sensing tasks are often annotated for objects (parcels of many pixels; e.g. a field), rather than being annotated at either image (classification) or pixel (segmentation) level. Differences such as these between satellite images and ground-based have encouraged researchers to explore novel data interpretations to use deep learning on satellite images in ways not typically seen in generic computer vision research. For example, a single satellite image might naturally be used in a 2DCNN, but might also be flattened and used in an MLP, or the spectral data might be considered a sequence and be used in a 3DCNN. To describe these differences and quantify their utilisation, we introduce a taxonomy of data interpretations in Section \ref{section:taxonomy}.

Large labelled datasets are critical for training robust deep learning models. But, while there is an abundance of images, the corresponding labels can be much more difficult to come by. For some tasks, like crop segmentation, the labels can be discerned directly from the image, but for most agricultural tasks, the target quantity is not so directly visible. This can be because the relationship between reflectance and the target value is complicated by various soil and biochemical attributes, as in the case of predicting Leaf Area Index (LAI), or because the target quantity is only knowable by analysing a time sequence, as in yield prediction. For such tasks, collecting data from ground level is necessary, but is more expensive to obtain. So, this review, analyses the data sources for each task and highlights the publicly available data.

The ultimate goal of most agricultural research is to help improve the yield and quality of our crops. But, the processes which turn sunlight, water, carbon dioxide, nutrients and minerals into the food we eat are varied and complex. They can manifest as broad visible changes, or as subtle chemical changes. Agricultural research can target any one of those pathways. By using a systematic search, this review identifies which agricultural quantities researchers are attempting to measure from satellite images using deep learning in practice (see Section \ref{section:tasks}). However, there remains open questions. Which quantities could or should be measured from space? Are there subtle signals in satellite images that truly provide information about the plants? Can deep learning uncover them if there are? While there have been some successes using deep learning on satellite images in crop segmentation and yield prediction, difficult challenges remain for other tasks.

In summary, the contributions of this review are:
\begin{enumerate}
    \item A gentle introduction to the use of satellite images, and how this differs to generic computer vision tasks.
    \item A taxonomy of data shapes and interpretations, and a quantification of how often each is used for each task.
    \item A tabulated list of references which includes this taxonomy, identifying which methods were used and which worked best in each study (see Supplementary materials).
    \item Quantitative analysis of the performance of various deep learning approaches on agricultural tasks.
    \item An investigation of what datasets and data sources are available. 
    \item Identification of the breadth of agricultural tasks using satellite images, including general information, specific challenges and suggestions to help adopt/improve deep learning for each task.
\end{enumerate}

\section{Search strategy}
\label{section:systematic-search}

\begin{table*}
    \footnotesize
    \centering
    \caption{The search terms used in the query for Clarivate's Web of Science. There is an "AND" between each top-level concept (i.e. [Deep Learning] AND [Satellite] AND [Agriculture]), and an "OR" between each term under that. The list of specific crops comes from the CDL \cite{usda_national_2022}. The search interface enforces a limit to the number of "All" search terms, so only the abstract and topic were searched for specific agriculture terms. The full list of agricultural terms is available in the supplementary materials section.}
    \renewcommand{\arraystretch}{1.1}
\begin{tabular}%
    {%
        >{\centering\arraybackslash}p{0.25\textwidth}||%
        >{\centering\arraybackslash}p{0.08\textwidth}||%
        >{\centering\arraybackslash}p{0.06\textwidth}|%
        >{\arraybackslash}p{0.33\textwidth}|%
        >{\centering\arraybackslash}p{0.14\textwidth}%
    }
\textbf{Deep Learning} & \textbf{Satellite} & \multicolumn{3}{c}{\textbf{Agriculture}} \\
\hline
\hline
All & All & All & \centering Abstract & Topic \\
\hline
Deep Learn*, CNN, RNN, LSTM, GRU, Transformer, Neural Network, Deep Belief Network, Autoencoder & 
Satellite &  
Farm, Agri*, Crop & 
Wheat, Corn, Maize, Orchard, Coffee, Vineyard, Soy, Rice, Cotton, Sorghum, Peanut*, Tobacco, Barley, Grain, Rye, Oat, Millet, Speltz, Canola, ... [+52 more]  &
*wheat, *flower*, *berries, *melon*, *berry
\end{tabular}
    \label{tbl:search-terms}
\end{table*}

To create an initial list of papers, we used a search query for Clarivate's Web of Science. To broadly find papers at the intersection of deep learning, satellite images and agriculture, we used both generic and specific terms for each (see Table \ref{tbl:search-terms}). For deep learning, this was specific algorithm names. For agriculture this was specific crop names from the Cropland Data Layer \cite{usda_national_2022}. The resultant tagged library of studies is available as supplementary materials.


The initial search yielded 770 studies. We performed an initial rapid pass through the collection of studies to filter out studies that were not at the intersection of deep learning, satellite imagery and agriculture, ultimately yielding 193 studies. The majority of these studies were for crop segmentation and yield prediction, thus, the studies for those tasks were further filtered as follows: 

\begin{itemize}
    \item 2020 and earlier: study is included if it has at least x citations on Google Scholar ($x=50$ for crop segmentation; $x=25$ for yield prediction)
    \item Jan 2021 - October 2022: all were included.
\end{itemize}

We did not include methods using UAV imagery because we were interested in methods for resolving the tension between object size and pixel size in satellite imagery. For crop segmentation studies (Section \ref{section:LULC}), we only include studies which used multiple agricultural classes. Soil monitoring studies (Section \ref{section:soil}) often only implied an agricultural significance, but, since soil has such a strong influence on agriculture, and relatively few studies, we include all found soil monitoring studies, even if they did not explicitly have an agricultural motivation. 

Although this review is systematic, it is not exhaustive, and not just because of the above filtering. By limiting the review to studies indexed by Clarivate's Web of Science, we are deliberately selecting for higher profile works than if we included searches across all published literature. We rely on the manual filtering stage to ensure that we only include relevant works. And while the search terms may not reveal all possible relevant studies, we believe that they are sufficient to return a representative sample of all relevant studies. 

There was also some inconsistency in terminology in the reviewed studies. In the interest of clarity, and to assist anyone unfamiliar with these terms, the variations are summarised in Table \ref{tbl:terminology}.

\begin{table*}
    \footnotesize
    \centering
    \caption{Some definitions for (sometimes inconsistent) terminology found in the literature.}
\begin{tabular}{p{0.3\textwidth}|p{0.6\textwidth}}
\textbf{Words} & \textbf{Idea} \\
\hline
Radiative transfer; reflectance; backscatter & Reflectance is the proportion of light which reflects off of a surface. This is a physical property of the surface, and can be measured in a laboratory. Radiative transfer models describe the physical process of reflectance, while backscatter is reflectance that is the result of artificial lighting, typically microwaves. \\
\hline
Sub-pixel fractional estimation; Linear Unmixing Model; Linear Mixture Model & A model of a pixel as being some proportion of just a few types of land cover, and thus every pixel's colour is explainable as an (often linear) combination of these cover types (see Section \ref{section:satellite-images}) \\
\hline
Downscale; upsample; finer resolution & Downscaling and upsampling have the same meaning because there is a conflict in terminology between remote sensing scientists and computer scientists, with inverse meanings. In this review, we have used ``coarser'' or ``finer'' to avoid confusion. \\
\hline
Multitemporal images; time series; Satellite Image Time Series (SITS); temporal data & Indicates the use of temporal data. Generally, stacked images of the same location over weeks/months (see Section \ref{section:common-methods}) \\
\hline
Multi-layer perceptron (MLP); Artificial Neural Network (ANN); Deep Neural Network (DNN) & Although ANN can technically refer to any Neural Network, it is typically used to refer to a small MLP. Generally, DNN refers to an MLP, but a DCNN refers to a CNN specifically. \\
\hline
Model inversion & Training a statistical model to predict the inputs of a theoretical model from either ground-measured outputs, or outputs of the theoretical model itself. A good summary of the ways this is used is given in \cite{weiss_remote_2020}. \\
\hline
Object-based; field-based; parcel-based; superpixel & Using aggregated colour information across a whole object or field or parcel or superpixel for prediction. \\
\end{tabular}
    \label{tbl:terminology}
\end{table*}


\section{Satellite Images}
\label{section:satellite-images}

Objects imaged by satellites are typically significantly smaller than the ground spatial distance (GSD) covered by each pixel. For example, the colour of each pixel in a satellite image of farmland might be aggregated from hundreds, thousands or even millions of individual plants. This massive difference in scale between object and pixel sizes has encouraged researchers to focus on understanding the contents of individual pixels as a combination of various surface types. This naturally encouraged per-pixel algorithms \cite{blaschke_whats_2001, blaschke_geographic_2014}, rather than the typical computer vision approaches which primarily use the structured pattern of multiple spatially-related pixels to understand an image \cite{funck_image_2003, wang_weakly_2020}.

While the spatial resolution relative to the imaged objects is much worse for satellite imagery, the spectral resolution is often significantly better. Almost all satellite imagery have at least 4 colour channels (red, green, blue and near-infrared), many have more than 10 colour channels (e.g. Sentinel-2), and some have over 100 different colour channels \cite{song_predicting_2018}, providing significantly more information per pixel than typical ground-based sources. Additionally, satellite image sensors are calibrated to obtain functions for converting from sensor brightness to reflectance - a physical property of the imaged surface - which allows quantitative analysis of the Earth's surface which is (mostly) independent of illumination and sensor.

At its core, reflectance is simply a ratio between reflected and incident light 

\begin{equation}
\rho = \frac{r}{i}
\end{equation}

But determining each of these values is confounded by complex shape geometries, atmospheric effects, sensor calibration errors, unexpected solar variation and the stochastic nature of photons.

Theoretically, with sufficiently precise measurements all surfaces could be uniquely identified by matching each pixel to a spectral signature measured in a lab. Indeed, this ideal is the basis of many hand-crafted models (e.g. Linear Mixture Model \cite{adams_classification_1995}). But, such a precise sensor does not exist, and significant noise is introduced by the lack of spatial and spectral resolutions, on top of the above errors calculating reflectance. These significant sources of noise have lead to the dominance of machine learning algorithms to learn the varied appearances of surfaces from data \cite{lu_survey_2007, khatami_meta-analysis_2016}. Such machine learning algorithms require many training examples to discover the existence/degree of a relationship between reflectance and the variable of interest.

Fortunately, there are several sources of freely available satellite imagery with worldwide coverage to train these algorithms. In the reviewed studies, the most common were: 
\begin{itemize}
    \item Moderate Resolution Imaging Spectroradiometer \\(MODIS) imagery at 250-1000m resolution which has been publicly available since 2000, along with many model-based maps, such as land surface temperatures, evapotranspiration and leaf area index (LAI).
    \item Landsat imagery which has been freely available to the public since 2008 \cite{zhu_benefits_2019}, of which most reviewed studies used Landsat-8 imagery at 30m resolution.
    \item Sentinel imagery from the Sentinel program of the European Space Agency which has provided optical imagery at 10-60m resolution and Synthetic Aperture Radar (SAR) imagery at 5-40m resolution since 2014.
\end{itemize}

The resolution of these data sources can dictate the resolution at which analysis can be performed; for example, county-level yield prediction always uses MODIS imagery and field-level yield prediction always uses Landsat/Sentinel imagery. These are obvious pairings because MODIS pixels are larger than individual fields, and images of entire counties using Landsat/Sentinel imagery would require a significant amount of disk space and computation time. At the coarser resolutions, there was a strong preference in the reviewed articles to pose the problem as just time series analysis, rather than a spatio-temporal one. Further, in several works \cite{jin_extraction_2021, shelestov_exploring_2017, ndikumana_deep_2018} and datasets like LUCAS \cite{dandrimont_lucas_2021}, the problem is posed as a single-pixel problem, only providing labels for a set of sparsely distributed points. Although this doesn't preclude the use of CNNs \cite{ji_3d_2018}, such a dataset discourages it.

Spatial resolution in satellite imagery has increased over the years, such that some commercial satellite providers now sell images with resolution as fine as 34cm per pixel, a resolution sufficiently fine to detect individual trees from satellite images \cite{gomez_use_2010, li_large-scale_2019, ferreira_accurate_2021, lin_toward_2021}. This increased resolution has encouraged satellite imagery analysis to utilise the spatial information - as in generic computer vision - as well as the higher spectral resolution and reflectance calibration typically used for satellite images (e.g. \cite{engen_farm-scale_2021, sagan_field-scale_2021, waldner_detect_2021}). We note that although spatio-temporal input has the richest information, it is not always available. For example, very-high resolution commercial satellite imagery is expensive and sparsely collected, thus most studies using commercial satellite imagery operated on a relatively small number of individual images (e.g. \cite{rahman_exploring_2018, choung_comparison_2021, saralioglu_semantic_2022}).


\section{Deep Learning}
\label{section:deep-learning}


In many domains, machine learning has found accurate relationships in spite of many variations in appearance and much noise. The data-driven nature of machine learning techniques handles such variations and models arbitrarily complex relationships while simultaneously including tools to prevent overfitting to the noise. We found that in single-pixel problems, Random Forests (RFs), Support Vector Machines (SVMs) and Multi-layer Perceptrons (MLPs) were generally close competitors, with each method being more accurate in different studies in roughly equal proportions (e.g. \cite{feng_machine_2019, ju_optimal_2021, shelestov_exploring_2017}).

Compared to other machine learning methods, deep learning is known to be able to construct significantly more complex models \cite{goodfellow_deep_2016}, allowing them to be more robust to noisy training data. Additionally, deep learning models learn to create their own features, which greatly reduces the need for manual feature engineering. This comes at the price of requiring larger datasets to observe this improved performance. In all studies reviewed in all tasks except yield prediction, modern deep learning methods outperformed traditional machine learning methods. In yield prediction, 2DCNNs consistently outperformed traditional machine learning methods, but LSTMs did not.



In the literature, various algorithms are called ``deep learning''. In this review, we refer to three main types of modern deep learning: CNNs, RNNs and Attention. With a decade since AlexNet \cite{krizhevsky_imagenet_2012}, and an explosion of research, Convolutional Neural Networks (CNNs) are the current de facto standard in generic computer vision tasks. Recurrent Neural Networks (RNNs) are a common deep learning method for sequence modelling, but almost all studies use an extension of RNNs; either Long Short-term Memory (LSTM) or Gated Recurrent Unit (GRU). ``Attention'' can mean many different things; here we will use it to mean, specifically, multi-head attention as described by \cite{vaswani_attention_2017}, as this is the basis for the recently popularised Vision Transformers \cite{dosovitskiy_image_2021} which have outperformed CNNs on recent ImageNet competitions. In this review we use ``deep learning'' to mean any neural network method, and ``modern deep learning'' to exclude MLP-only algorithms. We will not discuss the technical details of these deep learning algorithms in this review; instead we will mention the most significant modern advances and refer the reader to existing explanations for more details \cite{goodfellow_deep_2016, brodrick_uncovering_2019, kattenborn_review_2021}.

The ImageNet classification dataset \cite{deng_imagenet_2009} has had an enormous influence on the trajectory of computer vision research. It is common - when deep learning is applied to a new domain - for authors to use architectures that achieved a high rank in the ImageNet competition. In particular, AlexNet \cite{krizhevsky_imagenet_2012}, VGG \cite{simonyan_very_2015} and ResNet \cite{he_identity_2016} have received the most attention. Similarly for segmentation models, models that performed well on the MS COCO and PASCALVOC datasets have been adopted. The most popular segmentation architectures are based on UNet \cite{ronneberger_u-net_2015} and DeepLabv3 \cite{chen_rethinking_2017}.

Two of the most significant innovations of modern deep learning are focused around training deeper models: skip connections \cite{he_identity_2016} and inter-layer normalisation (e.g. BatchNorm \cite{ioffe_batch_2015}, LayerNorm \cite{ba_layer_2016}, etc). These two ideas have been almost universally adopted by all popular modern deep learning architectures, and with modern programming libraries these are easily incorporated into custom architectures created by individual studies (e.g. \cite{benedetti_m3fusion_2018, gallo_sentinel_2021}). Many works reviewed also used Dropout \cite{hinton_improving_2012}; another popular addition to training deep learning models for training more robust models.

\section{Common methods}
\label{section:common-methods}

\subsection{Taxonomy}
\label{section:taxonomy}

\begin{figure}
    \centering
    \includegraphics[width=0.48\textwidth]{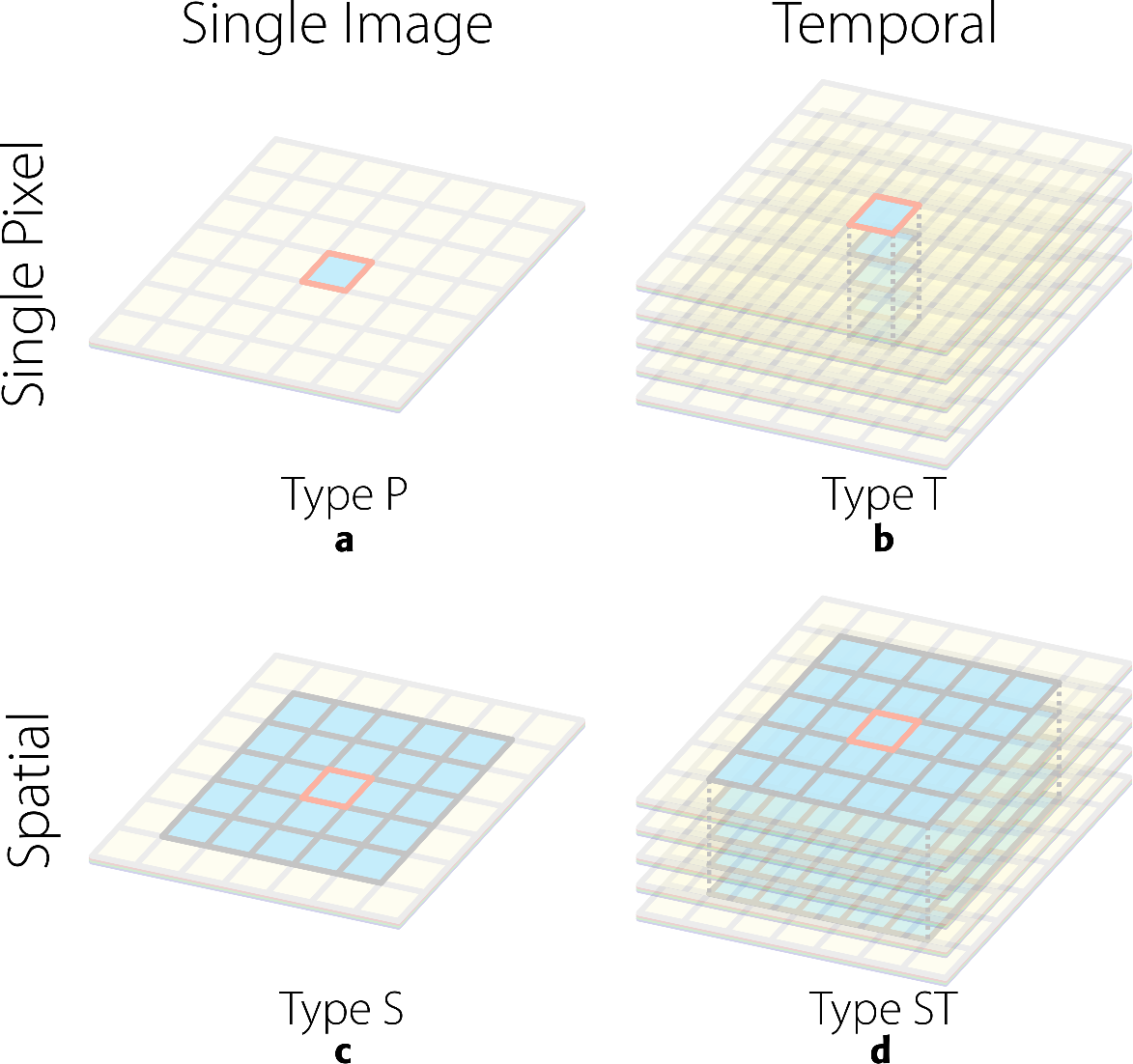} 
    \caption{Depending on the images available, a problem can be posed as: \textbf{a}) a relationship between a single pixel input (blue cell) from a single image (yellow grid of cells) and each prediction (red cell), or it can include contextual pixels from spatial (\textbf{b}) or temporal domains (\textbf{c}) or both (\textbf{d}), known as spatio-temporal (ST) data. For example, a model which operates on a sequence of co-located Sentinel-2 images would be said to use spatio-temporal input data.}
    \label{fig:data-cube}
\end{figure}

\begin{figure*}
    \centering
    \begin{subfigure}[t]{0.47\textwidth}
        \includegraphics[width=\textwidth]{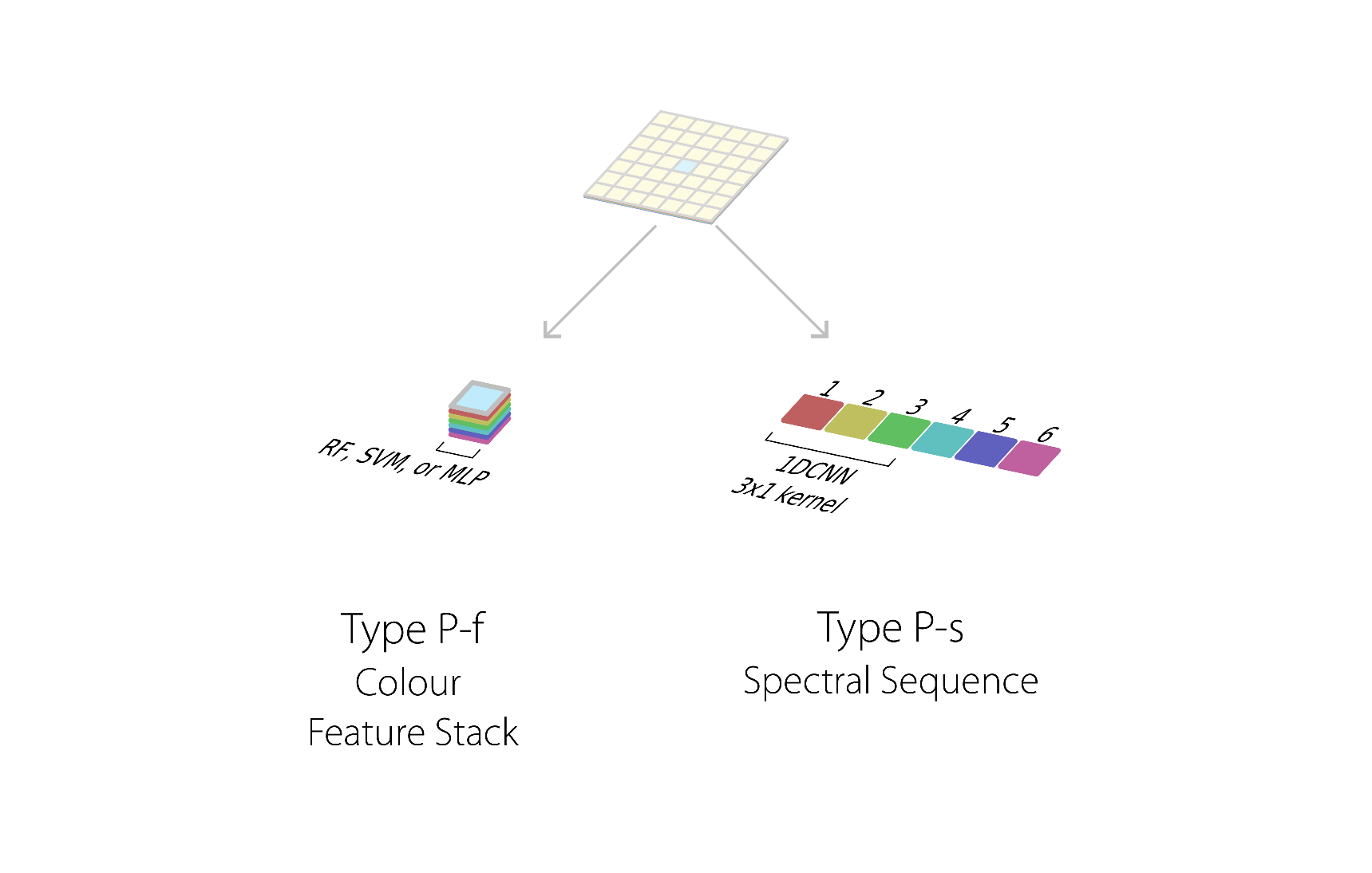}
        \caption{Type P (pixel data) can be interpreted as either a feature stack (P-f) or as a spectral sequence (P-s). For example, using a single pixel from a Landsat image.}
    \end{subfigure}
    \hspace{0.03\textwidth}
    \begin{subfigure}[t]{0.47\textwidth}
        \includegraphics[width=\textwidth]{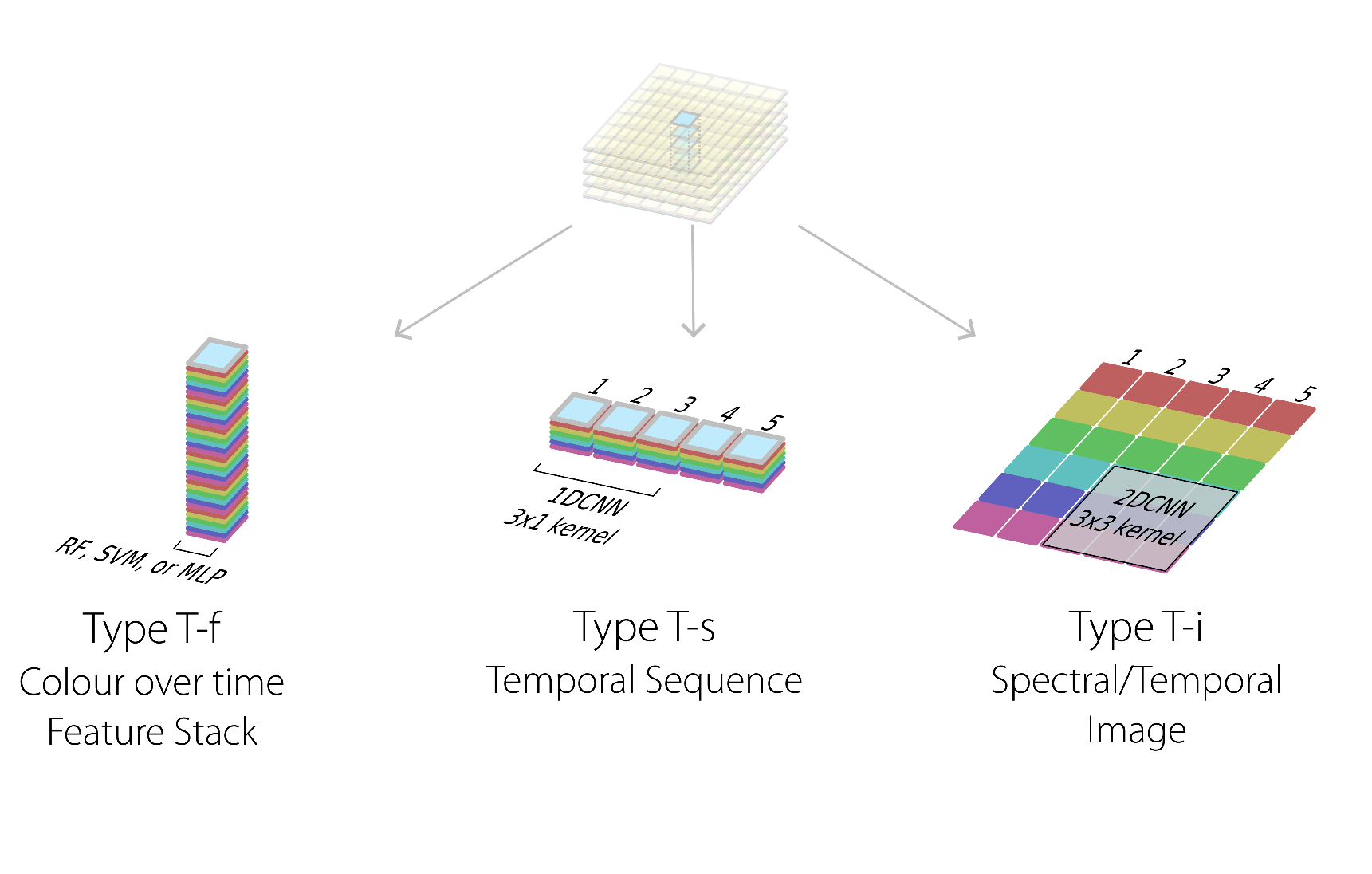}
        \caption{Type T (temporal data) can be interpreted as a single feature stack (T-f), as a sequence of feature stacks (T-s) or as an ``image'' with spectral and temporal dimensions (T-i).}
    \end{subfigure}
    \begin{subfigure}[t]{0.47\textwidth}
        \includegraphics[width=\textwidth]{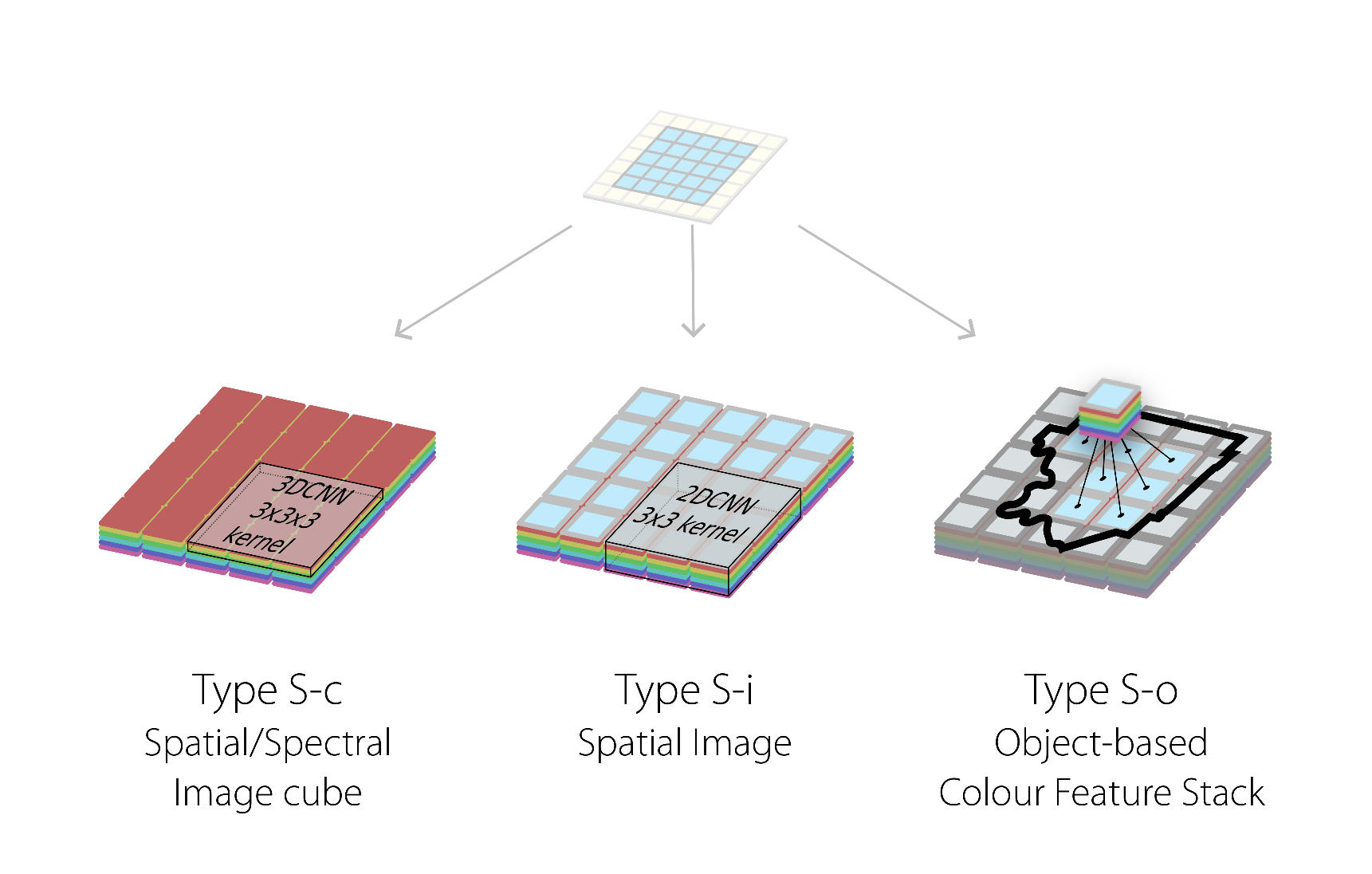}
        \caption{Type S (spatial data) can be interpreted as a ``data cube'' with two spatial and one spectral dimensions (S-c), as an image with just two spatial dimensions (S-i), or as a bundle of unstructured pixels that fit within an object boundary (S-o)}
    \end{subfigure}
    \hspace{0.03\textwidth}
    \begin{subfigure}[t]{0.47\textwidth}
        \includegraphics[width=\textwidth]{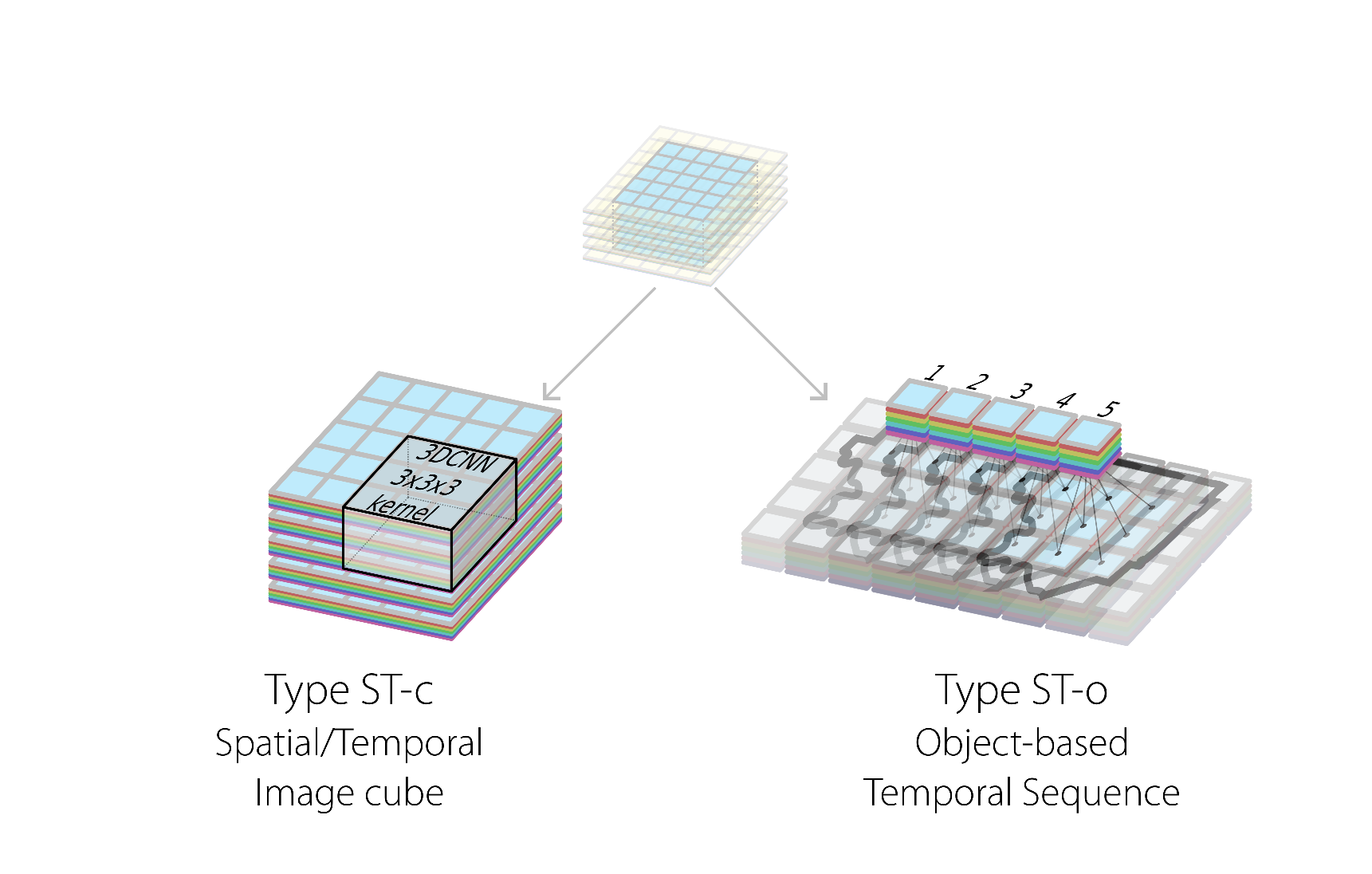}
        \caption{Type ST (spatio-temporal data) can be interpreted as a data cube with two spatial and one temporal dimensions (ST-c), or as temporal sequence of pixels within an object boundary (ST-o)}
    \end{subfigure}
    \caption{The data can be interpreted in different ways to allow the use of different models. Each of these data shape/interpretations pairs is given a name like type X-x to denote the original type and its interpretation. Subfigures here match those from Figure \ref{fig:data-cube}. i.e. \textbf{a} here show the interpretations of Figure \ref{fig:data-cube}\textbf{a}. The most notable distinctions are how spectral data is interpreted; as either a bag of features (P-f, T-s, S-i), or as a dimension in its own right (P-s, T-i, S-c). Although these depictions show spectral information, several studies replaced the spectral information with a set of other features: vegetation indices, topographical, atmospheric, soil, etc.}
    \label{fig:data-interpretations}
\end{figure*}

Satellite images are quantised measurements of our real world along multiple dimensions: one spectral, two spatial and one temporal. In the best case every prediction is based on a 4-dimensional data cube of spatio-temporal (ST) data, but such data can be computationally expensive to use or impractical to obtain, so many works operate on data without a spatial or temporal (or both) dimension. The shape of input data puts important emphasis and limitations on models trained on such data: for example, LSTMs are usually applied by processing each temporal sequence of pixels independently, which can observe changes over time, but wouldn't be able to use spatial contextual clues in its prediction. Thus, we create a taxonomy of input types to help understand how satellite images are being used differently across studies and agricultural tasks. 

Typically, studies will describe their input data, but not how they interpret that data for models. For example, a model might be called "pixel-based" because it operates on one pixel at a time. However, there are actually two ways that a single pixel in remote sensing can be interpreted. Is it a bag of features, or is it a spectral sequence? The latter interpretation is relatively common, but there has not previously been any consistent language or system to identify this distinction. Thus we start from the typical terminology of pixel-based (type P), spatial (type S), temporal (type T) and spatiotemporal (type ST) (see Figure\ref{fig:data-cube}). Then, in Figure \ref{fig:data-interpretations} we describe our taxonomy of different interpretations of those data shapes which authors have used to structure their data for use in modern deep learning algorithms. 

We name the interpretations by their initial data shape as the first character and their interpreted shape as the second character (i.e. [Shape]-[Interpretation]). The initial data shapes can then be interpreted as one of: a vector/bag of features (type X-f), a sequence of pixels (type X-s), a single image (type X-i), an image cube (type X-c), or an object (type X-o). For example, type T-s denotes an interpretation of a temporal sequence of pixels (T) for models that take advantage of the fact that it is a sequence (s), while type T-f denotes an interpretation of temporal data (T) as an stack of features with no time dimension (f) (see Table \ref{tbl:taxonomy-examples} for more examples). In Section \ref{section:tasks}, we use this taxonomy to quantify the most frequent types used for each task, but this taxonomy could also be useful outside of this review.

\begin{table*}
\centering
\caption{The taxonomy type names are short-hand for sentences to facilitate easier communication of relatively subtle distinctions. Here are the most common types encountered.}
\begin{tabular}{p{0.25\textwidth}p{0.07\textwidth}p{0.6\textwidth}}
\textbf{Name} & \textbf{Type} & \textbf{Description} \\
\hline
Pixel feature stack & P-f & A pixel-based algorithm that interprets the colour channels in a pixel as a single feature stack. \\
Pixel spectral sequence & P-s & A pixel-based algorithm that interprets the colour channels in a pixel as a spectral sequence. \\
\hline
Colour over time feature stack & T-f & An algorithm that interprets a sequence of pixels as a single feature stack \\
Temporal sequence & T-s & An algorithm that uses a temporal sequence of pixels and takes advantage of that fact. \\
Spectral/Temporal Image & T-i & An algorithm that interprets a temporal sequence of pixels as an image by treating the spectral dimension the same as a spatial one. \\
\hline
Spatial/Spectral Image Cube & S-c & An image-based algorithm that treats the spectral part of spatial data as a sequence to create an image cube. \\
Spatial image & S-i & An algorithm that uses a spatial image and takes advantage of that fact. \\
Spatial feature stack & S-f & An image-based algorithm that interprets spatial data as an unstructured feature stack. \\
Object-based pixel spectral sequence & S-o(P-s) & An algorithm that uses aggregated spatial data to form an average pixel and then interprets that pixel as a spectral sequence. \\
Object-based pixel feature stack & S-o(P-f) & An algorithm that uses aggregated spatial data to form an average pixel and then interprets that pixel as a spectral sequence. \\
\hline
Spatio-temporal image stack & ST-i & An algorithm that uses spatio-temporal data which interprets that 4D data cube as a non-temporal image with extra colour channels. \\
Spatial/Temporal image cube & ST-c & An algorithm that treat spatio-temporal data as a 4D data cube, but treating the spectral dimension as a bag of features. \\
Object-based temporal feature stack & ST-o(T-f) & An algorithm that uses spatially aggregated data to form a temporal sequence and then interprets that sequence as an unstructured feature stack. \\
Object-based temporal sequence & ST-o(T-s) & An algorithm that uses spatially aggregated data to form a temporal sequence and then takes advantage of it as a sequence. \\
Object-based pixel feature stack & ST-o(P-f) & An algorithm that uses data aggregated across both space and time at the same time to produce a single feature stack with no explicit time dimension. \\
Object-based temporal data cube & ST-o(T-c) & An algorithm that aggregates across space as a histogram of multiple statistics to create a 3D data cube, treating the histogram bins and the different statistics as ordered dimensions the same as the time dimension. \\
\hline
\end{tabular}

\label{tbl:taxonomy-examples}
\end{table*}

Traditional machine learning methods such as Random Forests (RFs), Support Vector Machines (SVMs) and Multi-layer Perceptrons (MLPs) generally require a 1D vector of features as input (types P-f and T-f), but modern deep learning methods generally have a multi-dimensional / sequential interpretation of the data (see Table \ref{tbl:model-method}). Several papers reframed the data to make specific modern deep learning methods more obvious candidates. The most common reinterpreting observed was treating spectral information as a sequence in order to apply sequence analysis algorithms. For example, type P-f and type P-s have precisely the same data, but by interpreting a single pixel (type P) as a sequence of colours (type P-s), 1DCNNs and LSTMs become obvious model choices. Similarly, temporal data (type T) can be interpreted as a spectral/temporal ``image'' (type T-i), and spatial data (type S) can be interpreted as a spatial/spectral image cube (type S-c) to use higher dimensional CNN kernels. Interpreting spectral information as a sequence is rarely theoretically justified, but multiple authors noted improved performance doing so \cite{kussul_deep_2017, debella-gilo_mapping_2021, saralioglu_semantic_2022}.

Sometimes the interpretation comes about because of labels. In yield prediction especially (see Section \ref{section:yield}), ground labels are often annotated at a field or county or country level (collectively, ``objects'') which are shown as types S-o and ST-o, depending on whether temporal data was included. In such cases, the relationship between the colour in the pixels and the final prediction are not necessarily present in each and every pixel contained within the the object, so a common method is to aggregate the information within an object before a model sees the data. Most commonly, the colour information is averaged, however \cite{khaki_simultaneous_2021} showed good performance by computing a colour histogram per county instead. These interpretations remove the spatial component, so after aggregation, S-o is subsequently re-interpreted as either P-f or P-s, and ST-o is subsequently re-interpreted as one of T-f, T-s or T-i. 

It is worth noting that a single 2D image is actually a three-dimensional tensor of data: two spatial and one spectral. A 2DCNN operates on this 3D data by treating the spectral information as the ``channels'' dimension. Similarly, a temporal stack of 2D images is a four-dimensional tensor of data: two spatial, one spectral and one temporal. A 3DCNN operates on a 4D ``image cube''. Thus, in the literature, an ``image cube'' may refer to the 3D data from a single 2D image after adding a singleton ``channels'' dimension (type S-c), or it may refer to the standard 4D data, using the spectral information as the ``channels'' dimension (type ST-c).

The terminology used for this taxonomy can also be used for combinations not shown in Figure \ref{fig:data-interpretations}. For example, \citet{chelali_deep-star_2021} and \citet{paul_generating_2022} both interpreted spatiotemporal as an image (type ST-i) without aggregation and in completely different ways. This taxonomy does not exhaustively describe every possible way to use data with modern deep learning algorithms, but describes the main cases with a consistent terminology. This enables more effective communication of technical details of algorithms.


\begin{table*}
\centering
\caption{Each model has a required input shape, and each data interpretation is intended to re-arrange the input to fit the model. Type X-f is any feature stack interpretation, type X-s is any sequence interpretation, type X-i is any image interpretation and type X-c is any image cube interpretation (see Figure \ref{fig:data-interpretations}).}
\newcommand{\rot}{\rotatebox{90}}

\begin{tabular}{c|c|c|c|c|c|c|c|c|c|c}

 & \rot{Tree} & \rot{SVM} & \rot{MLP} & \rot{1DCNN} & \rot{2DCNN} & \rot{3DCNN} & \rot{RNN} & \rot{CNN + RNN} & \rot{ConvRNN\cite{shi_convolutional_2015}} & \rot{Attention} \\
\hline
Type X-f & \checkmark & \checkmark & \checkmark &  &  &  &  & &  & \\
\hline
Type X-s &  &  &  & \checkmark &  &  & \checkmark & & & \checkmark \\
\hline
Type X-i &  &  &  &  & \checkmark &  &  &  & & \checkmark \\
\hline
Type X-c &  &  &  &  &  & \checkmark &  & \checkmark & \checkmark & \checkmark \\

\end{tabular}
\label{tbl:model-method}
\end{table*}

\subsection{Clouds}

Cloud cover and other atmospheric effects present significant challenges to Earth Observation tasks using optical imagery by obscuring the targets of interest. It is common to simply exclude single and multitemporal pixels that are occluded by clouds \cite{kussul_deep_2017} when operating at a pixel level (type P or type T), or interpolate the colour information from earlier and later images (e.g. \cite{inglada_assessment_2015, benedetti_m3fusion_2018, interdonato_duplo_2019}) when using temporal data (type T or ST). Another common strategy is to create single image composites from images taken at different times (e.g. \cite{waldner_deep_2019, lu_mapping_2022}). Some authors remove entire images from their training set if there is too much cloud cover \cite{moreno-revelo_enhanced_2021}. Discarding entire samples reduces the number of examples available for model training, and when using multitemporal data, discarding either the whole multitemporal pixel or individual images in the sequence degrades performance. One solution used by \citet{rawat_deep_2021} was to collect both Landsat-8 and Sentinel-2 images, and when a Sentinel-2 image was too cloudy, replace it with a Landsat-8 image. And \citet{metzger_crop_2022} show a method to explicitly encode variable temporal sampling into a recurrent model's architecture to prevent performance degradation. Other authors argue that deep learning is robust to occasional entire images being completely covered in cloud and use the data without explicitly handling clouds \cite{engen_farm-scale_2021, turkoglu_crop_2021, ruswurm_convolutional_2018}.

Cloud cover poses no issue for Synthetic Aperture Radar (SAR) imagery, so cloud cover can be mitigated by using both optical and SAR imagery (e.g. \cite{wang_mapping_2020}). Few works took advantage of Sentinel-1 SAR imagery in addition to their own, but those that did noted improved performance \cite{wang_mapping_2020, moumni_machine_2021, thorp_deep_2021, ofori-ampofo_crop_2021}.

\subsection{Temporal mismatch}

Ground-based in situ measurements are generally not coordinated with satellite imagery. Instead the measurements are taken, and then paired with images at the closest available time, leading to a temporal mismatch. In the reviewed studies there are three methods for handling this temporal mismatch (roughly in order of popularity): ignore it; interpolate ground labels to match image collection date \cite{kira_toward_2017}; or interpolate images to match measurement date \cite{bocco_estimating_2012}. Ignoring the mismatch is popular because it is a reasonable action to take if the property is not expected to change significantly during the timescale of mismatch, and most properties of interest do not change much over a few days to a week.

\section{Tasks}
\label{section:tasks}

\begin{table*}
    \footnotesize
    \caption{Number of papers for each task. By far, Crop segmentation was the most popular, followed by County-level yield prediction.}
    \centering
    \begin{tabular}{lllll}
    \textbf{Land Use, Land Cover} & \textbf{Soil Monitoring} & \textbf{Plant Growth} & \textbf{Disease/Damage} & \textbf{Yield Est.} \\
    \hline 
    Crop Segmentation (73) & Moisture (14) & Canopy cover/LAI (14) & Disease (6) & County-level (23) \\
    Field Boundary (8) & Nutrients (6) & Growth Stage (2) & Damage (4) & Field-level (7) \\
    Tree Crown Delineation (6) & Salinity (4) & Other (3) & & Plot-level (2) \\
    Dam Detection (4) & & & & Pixel-level (2) \\
    Other (8) & & & & Other (7) \\
\end{tabular}
    \label{tbl:tasks}
\end{table*}

The reviewed studies were manually categorised into one of 5 groups of tasks (see Table \ref{tbl:tasks}). These categories were determined based on the papers reviewed, rather than being imposed from the beginning and are roughly arranged in order of the plant's life cycle: first, we consider how deep learning is used to describe where crops are being grown and the location of other farm equipment/landmarks (Section \ref{section:LULC}). Next, we explore how deep learning is used to evaluate the soil's health (Section \ref{section:soil}). We then consider methods for observing plants after establishment: evaluating the plant's growth (Section \ref{section:physiology}), monitoring damage from disease or disaster (Section \ref{section:damage}), and finally estimating crop yield (Section \ref{section:yield}). See supplementary materials for the full list of studies.

\subsection{Land Use and Land Cover}
\label{section:LULC}

\begin{figure*}
    \centering
    \includegraphics[width=0.8\textwidth]{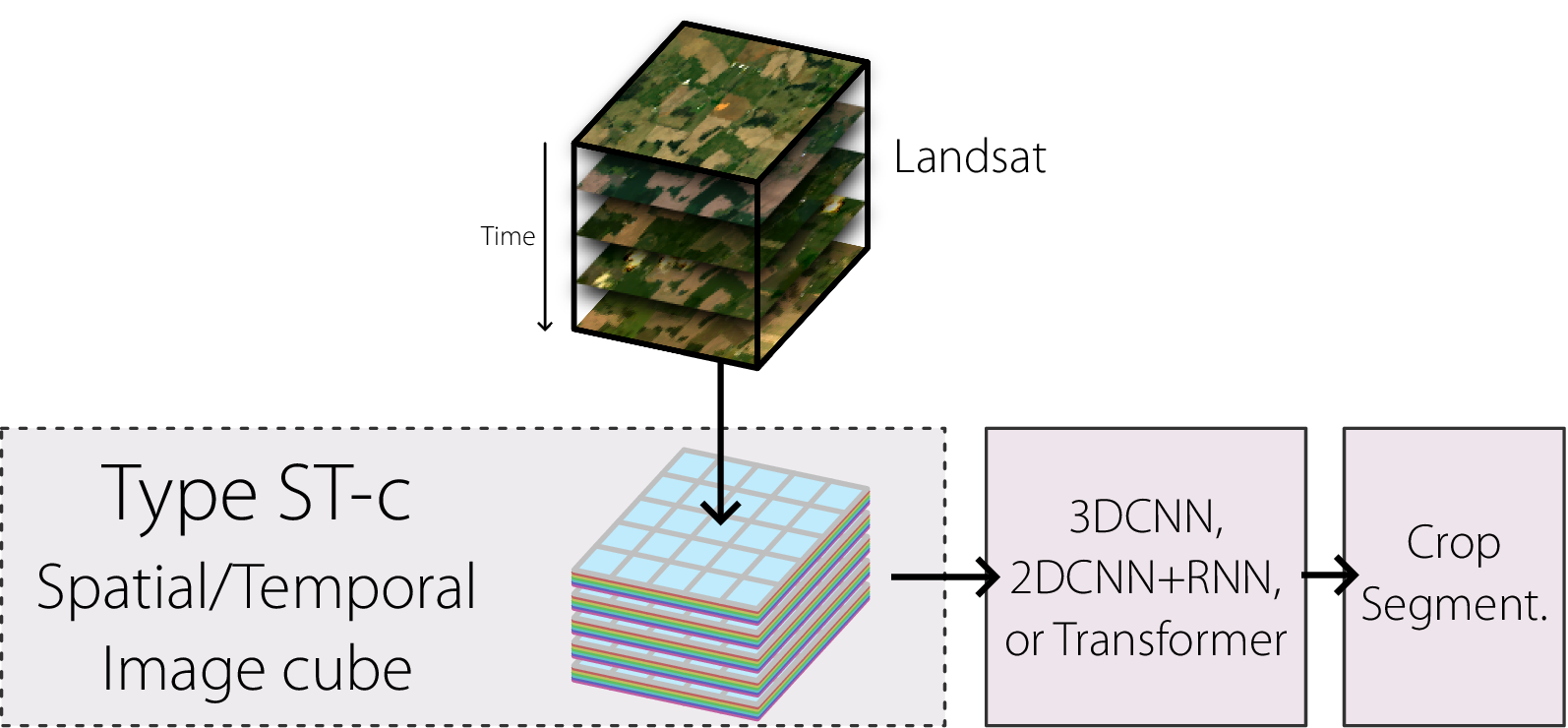}
    \caption{One of the more common input methods for Crop Segmentation was spatiotemporal data treated as a 4D image cube (type ST-c). This type of data typically comes from Landsat or Sentinel data sources which provide images at a uniform temporal resolution, and can be used in a 3DCNN, or 2DCNN/RNN hybrid, or Transformer models.}
    \label{fig:type-stc}
\end{figure*}

\begin{table*}
    \footnotesize
    \caption{Counts of studies on Land Cover and Land Use (LULC). VI counts studies that used vegetation indices. In the labels column: FS = Field survey, IS = Image survey, MB = Model-based, Govt. = Government. Note that studies may use multiple Models and types in the same study. Some uncommon types not shown; for full list see supplementary materials.}
    \centering
    \renewcommand{\arraystretch}{1.2}
    \newcommand{\rot}{\rotatebox{90}}
\newcolumntype{C}[1]{>{\centering\let\newline\\\arraybackslash\hspace{0pt}}m{#1}}
\newcolumntype{R}[1]{>{\raggedleft\let\newline\\\arraybackslash\hspace{1pt}}p{#1}}

\begin{tabular}{rc|c|ccccc|cccccc|ccccccccc} 
& \textbf{n} & \textbf{VI} & \multicolumn{5}{c|}{\textbf{Labels}} & \multicolumn{6}{c|}{\textbf{Model}} & \multicolumn{9}{c}{\textbf{Type}} \\
& & & \rot{FS} & \rot{IS} & \rot{MB} & \rot{Govt.} & \rot{Dataset} & \rot{Tree} & \rot{SVM} & \rot{MLP} & \rot{CNN} & \rot{RNN} & \rot{Attn} & \rot{P-f} & \rot{P-s} & \rot{T-f} & \rot{T-s} & \rot{S-i} & \rot{ST-c} & \rot{S-o(P-f)} & \rot{ST-o(T-f)} & \rot{ST-o(T-s)} \\
\hline
\textbf{Crop Seg.}      & 73 & 25 & 26 & 17 & 8 & 18 & 10 & 28 & 21 & 23 & 54 & 31 & 14 & 17 & 3 & 12 & 18 & 18 & 18 & 2 & 4 & 7 \\
\textbf{Field Bounds} &  8 &  0 &  0 &  7 & 0 &  2 &  0 &  0 &  0 &  0 &  8 &  0 &  0 &  0 & 0 &  0 &  0 &  8 &  0 & 0 & 0 & 0 \\
\textbf{Dam Det.}  &  4 &  0 &  0 &  3 & 0 &  0 &  0 &  0 &  1 &  0 &  3 &  0 &  0 &  1 & 0 &  0 &  0 &  3 &  0 & 0 & 0 & 0 \\
\textbf{Tree Bounds}     &  6 &  1 &  2 &  4 & 0 &  0 &  0 &  0 &  0 &  2 &  3 &  1 &  1 &  2 & 0 &  0 &  0 &  4 &  0 & 0 & 0 & 0 \\
\textbf{Other}          &  8 &  0 &  2 &  7 & 0 &  1 &  0 &  0 &  0 &  0 &  8 &  0 &  0 &  0 & 0 &  0 &  0 &  8 &  0 & 0 & 0 & 0 \\
\end{tabular} 
    \label{tbl:agg-lulc}
\end{table*}

The task of Land Use and Land Cover (LULC) is to either classify each pixel of satellite/UAV imagery (segment), or detect specific regions/objects in the images. LULC is by far the most common use of deep learning in satellite imagery in the reviewed articles (98/193), and within LULC crop segmentation was the most popular subtask (73/98 LULC studies; see Table \ref{tbl:agg-lulc} and supplementary materials). This popularity is not surprising, as land cover mapping is considered the most important descriptor of the environment \cite{herold_evolving_2006}. The power of LULC is widely recognised and governments across the world publicly provide large-scale LULC maps (e.g. \cite{usda_national_2022, dandrimont_lucas_2021, abares_catchment_2021}), created through a combination of visual inspection of satellite imagery, census information, local expertise and automated methods (e.g. linear unmixing model). Automated methods are the least reliable of these, but are the easiest to scale up. Thus, the goal of research into LULC and crop segmentation is to improve the quality of the automated methods so that eventually they become at least as reliable as expert humans visually inspecting satellite images.

\subsubsection{Existing Approaches}

Out of all of the tasks reviewed in this paper, the deep learning methods for crop segmentation were the most varied and novel, with researchers using most of the types from the taxonomy in Section \ref{section:taxonomy}, and more. There seems to be two main reasons for this: first is the popularity of the task, and second is the relative ease of obtaining data, allowing datasets large enough to train modern deep learning methods. Most works reviewed were using at least 2000 data points, and many were in the 10,000s, which - while small by deep learning standards - are significantly larger than datasets used for the other reviewed satellite tasks. All major modern deep learning algorithms are represented in the reviewed papers: 1DCNNs (19 studies), 2DCNNs (27), 3DCNNs (8), LSTMs (22), GRUs (9), ConvLSTM/ConvGRU (7) and transformers (8)/other attention (6), with many studies comparing between these methods. In the reviewed articles, the best modern deep learning methods (CNN/RNN/Attn) outperformed all tree (22 comparisons) (including RF (21 comparisons)), SVM (12 comparisons) and MLP (6 comparisons) methods in all cases. It was observed that in many studies there were several deep learning algorithms tested, and almost all of these outperformed trees, SVMs and MLPs as well. The LULC tasks other than crop segmentation were generally detection tasks, and generally followed generic computer vision: almost all of these studies used 2D spatial CNNs on individual satellite images (type S-i; 22/25).

The first use of modern deep learning techniques in the reviewed studies was \citet{kussul_deep_2017}, who used their own custom architectures with five convolution layers using a dataset of 100,000s of labelled pixels from field surveys and found that 2DCNNs outperformed RF, MLP and 1DCNNs. From this time onwards, the majority of the papers reviewed used modern deep learning (excluding purely MLP methods): there were 28 deep learning studies reviewed from 2021 and 21 of those used modern deep learning methods. 

Most studies in crop segmentation utilising 2DCNNs created their own architectures / arrangement of layers, typically favouring shallower networks (<10 layers), with only a few using common architectures used in generic computer vision like VGG, ResNet and UNet (e.g. \cite{ji_3d_2018, sagan_field-scale_2021, waldner_detect_2021}). However, in the other LULC tasks, using those existing architectures was extremely common, with all but two studies \cite{persello_delineation_2019, masoud_delineation_2020} using an existing architecture, and several using pretrained weights from ImageNet (e.g. \cite{li_agricultural_2020, ferreira_accurate_2021}). There were two main algorithms used to process spatiotemporal data (type ST-c; see Figure \ref{fig:type-stc}) the first is to use a 3DCNN (e.g. \cite{ji_3d_2018, wang_mapping_2020, gallo_sentinel_2021}), and the second is to use 2DCNNs on the images and then use RNNs (GRUs/LSTMs) on the outputs (e.g. \cite{ruswurm_self-attention_2020, li_adversarial_2021, teimouri_novel_2019}), however these methods were not compared against each other.

In generic computer vision, transformers \cite{dosovitskiy_image_2021} achieve state-of-the-art performance on some major benchmarks, so one might wonder about their performance on satellite tasks. There were 8 studies that directly referenced transformer networks, and a further 6 that used a similar multi-head attention mechanism as is used in transformers. In all works that compared a transformer network to other modern deep learning methods, the transformer performed equal or worse \cite{ruswurm_self-attention_2020, xu_deepcropmapping_2020, turkoglu_crop_2021, metzger_crop_2022, tang_channel_2022, sykas_sentinel-2_2022}. Although, we are aware of at least one study from outside the search criteria which shows an improvement\cite{kondmann_denethor_2021}. Regardless, this is surprising. Transformer-based models would be expected to generally perform better. It has been noted that typical transformer networks require even more training data than CNNs \cite{dosovitskiy_image_2021}, stating that even ImageNet (1 million labelled images) is too small, so it is possible that the datasets used in these satellite studies are simply not large enough for transformer networks to begin to outperform CNNs. The main other attention method used is a model called Pixel-Set Encoder and Temporal Attention Encoder (PSE-TAE) \cite{garnot_satellite_2020}. This was compared against other methods in only one of the reviewed studies, and it was out-performed by a 1DCNN with a Squeeze and Excitation module \cite{tang_channel_2022}.

\subsubsection{Datasets}

When creating data sources for developing LULC models it is impossible to both manually annotate every pixel and then manually verify each field at ground level. There is also inherent ambiguity of what is visible in these images, so, even with the huge manual effort that goes into creating these data sources, they are known to have significant error in them; e.g. the Cropland Data Layer (CDL) \cite{usda_national_2022} has an error rate of 5-15\% for major crop types \cite{boryan_monitoring_2011}. Many European countries require that farmers register what crops they grow - along with farm boundaries - when applying for government rebates. Notably, France, Switzerland, Norway and others have started to provide this data publicly in a quasi-anonymised form providing huge data sources of labels to train machine learning models, and is the basis of most public benchmark datasets for LULC. These initiatives create high quality data sources with few errors. Although it should be noted that farmers may accidentally or deliberately misreport the crops they are growing \cite{hamer_replacing_2021}, so even these are not error-free.

\begin{table*}
    \footnotesize
    \centering
    \caption{Identified crop segmentation satellite benchmark datasets and data sources. Benchmark datasets (those with a tick in B) have a clearly defined size, and thus can be used to compare different methods. Only the most popular public government (non-benchmark) sources are included here.}
    \newcommand{\rot}{\rotatebox{90}}

\begin{tabular}{>{\raggedright}p{80pt}|c|c|p{50pt}|>{\raggedleft}p{45pt}|p{75pt}|p{35pt}|c}
\textbf{Name} & \textbf{Labels} & \textbf{Task} & \textbf{Location} & \textbf{Years} & \textbf{Size} & \textbf{Source} & \textbf{B} \\
\hline
\rowcolor{gray!10} LUCAS \cite{dandrimont_lucas_2021} & Point Data & Classification & Europe & 2006-2018 & 63,000 points; 11.8MB & In Situ / Visual & \checkmark \\
Munich dataset \cite{ruswurm_multi-temporal_2018} & Shapefile & Semantic Seg. & Munich & 2016-2017 & 137,000 fields; 41.5GB & Govt. & \checkmark \\
\rowcolor{gray!10} BreizhCrops \cite{breizhcrops2020} & Superpixels & Classification & Brittany & 2017 & 750,000 superpixel sequences; 4.6GB & Govt. & \checkmark \\
DENETHOR \cite{kondmann_denethor_2021} & Shapefile & Semantic Seg. & Germany & 2018-2019 & 4,500 fields; 254.5GB & Govt. & \checkmark \\
\rowcolor{gray!10} ZueriCrop \cite{turkoglu_crop_2021} & Shapefile & Semantic Seg. & Switzerland & 2019 & 116,000 fields; 38.5GB & Govt. & \checkmark \\
Reunion Island \cite{dupuy_reunion_2020} & Shapefile & Semantic Seg. & Reunion Island & 2019 & 50,000 fields; 290MB & Model & \checkmark \\
\rowcolor{gray!10} Campo Verde \cite{sanches_campo_2017} & Shapefile & Semantic Seg. & Brazil & 2015-2016 & 500 fields; 6.7GB & Visual & \checkmark \\
UOS2 \cite{pedrayes_evaluation_2021} & Segmentation & Semantic Seg. & Spain & 2020 & 2,000x 256x256 tiles; 5.5GB & Visual & \checkmark \\
\rowcolor{gray!10} PASTIS-R \cite{garnot_multi-modal_2022} & Segmentation & Semantic Seg. & France & 2019 & 2,400x 128x128 tiles; 54GB & Govt. & \checkmark \\  
Sen4AgriNet \cite{sykas_sentinel-2_2022} & Segmentation & Semantic Seg. & France, \newline Catalonia & 2019-2020 & 250,000x S2 images; 10TB & Govt. & \checkmark \\ 
\rowcolor{gray!10} French Land Parcel Identification System & Shapefile & - & France & - & - & - &  \\
Cropland Data Layer \cite{usda_national_2022} & Segmentation & - & USA & - & - & - &  \\
\rowcolor{gray!10} Corine Land Cover & Segmentation & - & Europe & - & - & - &  \\
WorldCereal \cite{boogaard_worldcereal_2023} & Shapefile & - & World & 2017-2021 & 74 million fields & - & 

\end{tabular}
    \label{tbl:datasets}
\end{table*}

This review makes a distinction between a \emph{data source} and a \emph{benchmark dataset}. The former is a living collection of data that is frequently updated and covers wide areas, whereas a benchmark dataset is a fixed set of images and labels on which different methods can be compared directly across studies. LULC was the only task to have any benchmark dataset or named data sources. See Table \ref{tbl:datasets} for the list of data sources and benchmark datasets used by the reviewed studies.

Despite the availability of this data, only 22 out of the 73 reviewed crop segmentation studies used continually updated government data sources, and 9 used benchmark datasets. Of these, only 4 were on the same dataset as another study, and only 3 compared against other algorithms on the dataset. And among these, none were algorithms created specifically for that dataset; instead they were transferred from other, similar tasks. Over half of the crop segmentation studies (46/73) described methods for obtaining their own segmentation maps. There are three methods studies used to do this: manual field survey (26/73), manual image survey (17/73), and automated methods (8/73). Manual field surveys are the more difficult, expensive and reliable way to obtain ground data for training models. Using an automated method to obtain training data for models is cheaper and less reliable, and largely limits deep learning models to matching the automated method's dynamics, rather than allowing it to learn the true distribution, so a few works used a combination of each. For example, \citet{rahimi-ajdadi_remote_2021} gathered 55 data points from field surveys, 360 from visual inspection and 550 from automated methods. And \citet{zhou_long-short-term-memory-based_2019} and \citet{hamer_replacing_2021} used automated systems to initially label their satellite images, then manually fix the errors. Several works relied entirely on manual image surveys (e.g. \cite{ji_3d_2018, ndikumana_deep_2018}), however only \citet{saralioglu_semantic_2022} described a validation method for their manual image survey.

\subsubsection{Analysis}

Land Use and Land Cover has recently seen an explosion of data sources that, together, provide a viable base to produce sophisticated deep learning solutions. And these sources are being collected and collated into ever larger training datasets (e.g. \cite{van_tricht_worldcereal_2023}). As such, deep learning has been successfully and convincingly applied to LULC, incorporating the latest generic computer vision research \cite{garnot_panoptic_2021, ruswurm_self-attention_2020}. However, in the reviewed articles, almost all work was focused on creating static crop maps in similar locations. In the future, more work will be needed to model dynamic, multi-annual changes \cite{liu_mapping_2022}, and to enable extending these models to regions of the world with little training data \cite{van_tricht_worldcereal_2023}.

\subsubsection{Recommendations}

Out of the 73 crop segmentation studies reviewed, 10 used Landsat-8, 40 used Sentinel-2 and 18 used Sentinel-1. Thus, there is a strong preference in modern research to use Sentinel-2 over Landsat-8, likely due to the superior resolution. Additionally, six studies evaluated the benefit of using both Sentinel-1 (SAR; active microwave) and Sentinel-2 (passive optical imaging). They unanimously found that taking advantage of the complementary information contained in each improved performance\cite{wang_mapping_2020, moumni_machine_2021, thorp_deep_2021}. In particular, late fusion (averaging two model's outputs) was found to be the most effective way to combine images from the two sources \cite{garnot_multi-modal_2022}. So, where appropriate it is recommended to combine Sentinel-1 and Sentinel-2 images.

For crop segmentation, unless there is an identified need, it is recommended to use an existing dataset for better comparisions with state-of-the-art.

2DCNNs appear to dominate all LULC tasks. These should be the first consideration for baselines in future research.




\subsection{Soil health}
\label{section:soil}


\begin{table*}
    \footnotesize
    \caption{Counts of studies on Soil health. n is the total number of studies. VI counts studies that used vegetation indices. In the labels column: FS = Field survey, MBt/FSe = Model-based for train and Field Survey for evaluation, Self = Self-labelled, Govt. = Government. Note that studies may use multiple Models and types in the same study.}
    \centering
    \renewcommand{\arraystretch}{1.2}
    \newcommand{\rot}{\rotatebox{90}}

\begin{tabular}{rc|c|cccc|ccccc|cccc}
& \textbf{n} & \textbf{VI} & \multicolumn{4}{c|}{\textbf{Labels}} & \multicolumn{5}{c|}{\textbf{Model}} & \multicolumn{4}{c}{\textbf{Type}} \\
& & & \rot{FS} & \rot{MBt/FSe} & \rot{Self} & \rot{Govt.} & \rot{Tree} & \rot{SVM} & \rot{MLP} & \rot{CNN} & \rot{RNN} & \rot{P-f} & \rot{T-s} & \rot{S-i} & \rot{S-o(P-f)} \\
\hline
\textbf{Moisture}  & 14 & 6 & 6 & 5 & 2 & 1 & 3 & 2 & 13 & 1 & 1 & 12 & 1 & 1 & 1 \\
\textbf{Nutrients} &  6 & 2 & 6 & 0 & 0 & 0 & 4 & 2 &  6 & 0 & 0 &  4 & 0 & 0 & 1 \\
\textbf{Salinity}  &  4 & 3 & 4 & 0 & 0 & 0 & 1 & 2 &  3 & 1 & 0 &  3 & 0 & 1 & 0 \\
\end{tabular} 
    \label{tbl:agg-soil}
\end{table*}

Soil health has an impact on all surface vegetation, however it can be difficult to measure from space. This is because measuring changes from satellite images is limited to measuring properties which cause a visible spectral change. There were three types of papers identified by this survey: soil moisture, soil nutrients and soil salinity. While soil moisture may cause obvious changes in visible spectra, changes in soil nutrients and salinity cause much less significant spectral changes \cite{qi_soil_2020}. So, instead, they are often inferred by observing the effect on vegetation. At the ground level, soil health is typically measured using a probe inserted into the soil that measures the chemistry directly.

\subsubsection{Soil moisture}

Mapping soil moisture is useful for understanding hydrologic processes, vegetation states, and climatic conditions \cite{entekhabi_mutual_1996}. Active microwave imaging is often used to predict soil moisture because the dielectric properties of soil changes with moisture for microwaves, highlighting areas with high moisture more clearly than with optical images \cite{dobson_microwave_1985}. There are several satellite-based worldwide spatially and temporally continuous active microwave imaging satellites used to estimate soil moisture from space \cite{njoku_soil_2003, entekhabi_soil_2010, kerr_smos_2012, wagner_ascat_2013}. \citet{rabiei_method_2021} provide a summary of different benefits and drawbacks to using different remote sensing techniques. They point out that much existing work utilising the active microwave satellites and distributed networks of soil monitoring stations are at a much coarser resolution than is useful for a single field. In their work, they used relatively fine resolution optical imagery (10m) with CNNs (type S-i), which they found to be superior to other machine learning methods.

\subsubsection{Soil nutrients/heavy metals}
The correct balance of soil nutrients is essential for healthy plant growth, but this balance is not easily observable with remote sensing. In the presences of vegetation, soil nutrients can be correlated with remote sensing only when it causes a visual change. For example, a lack of nitrogen leading to poor plant growth. This means that such works on soil nutrients are limited to locations where such a pathway is the dominant one. And in the absence of vegetation, monitoring soil heavy metals is only feasible with exquisitely detailed spectral information \cite{rasheed_fluorescent_2018, zhang_retrieving_2022}. 

There were seven studies which attempted to measure soil nutrient or heavy metal content using deep learning on satellite imagery. All of them used MLPs, but only three of them compared with other methods. Within these, there was no consistent model which performed best: \citet{wang_exploring_2022} and \citet{zhang_retrieving_2022} found that RFs worked better than MLPs and \citet{song_predicting_2018} found that an MLP worked better than an RF or SVM. None of the works used CNNs or other modern deep learning architectures.

\subsubsection{Soil salinity}
High soil salinity is a growing problem worldwide, and there is growing interest in mapping this environmental hazard using satellite imagery \cite{metternicht_remote_2003}. Despite the dire impacts of high soil salinity, there were only 4 studies that investigated predicting soil salinity from satellite imagery using deep learning with reference to agriculture. Only one used a modern deep learning approach \cite{akca_semantic_2022}, and they found that it performed significantly better than SVM models using over 500 data points. The three other studies \cite{wang_estimation_2018, qi_soil_2020, habibi_quantitative_2021} only used MLPs and were small in scope, with less than 100 data points each.

\subsubsection{Existing Approaches}

In the reviewed works, there were only 24 using deep learning to measure soil properties with satellite images (see Table \ref{tbl:agg-soil}). Within this relatively small sample, the algorithms were remarkably consistent. Almost all of the works used a pixel-based algorithm (type P-f; see Figure \ref{fig:type-pf}): 22 of the 24 studies used an MLP, and of these, around half did not compare against any other method. There were three works that used modern deep learning methods: two used a 2DCNN (type S-i) \cite{rabiei_method_2021, akca_semantic_2022} and observed improved performance, and one used an LSTM on type T-s data \cite{zeynoddin_structural-optimized_2022} without comparing against other algorithms.

\subsubsection{Datasets}

Although there is at least one global soil moisture dataset \cite{dorigo_international_2011}, it does not provide homogeneous measurements across sites and is not well adopted in the reviewed studies. There are no publicly available soil nutrients or soil salinity datasets mentioned in the reviewed articles.

\subsubsection{Analysis}

MLPs are the dominant approach for deep learning on Soil health monitoring with satellite images. Soil measurements are generally point measurements, and this seems to encourage authors to use type P-f data, only. In other tasks, applying CNNs have improved results, but they have barely begun to be explored for soil health monitoring. This is likely because soil measurements are generally point samples, and it is not immediately obvious that it is possible to apply spatial models. Additionally, in spite of the observations that these values can change drastically over a season \cite{rabiei_method_2021}, only one work modelled soil properties over time. 

The reviewed works generally had very small sample sizes; only one used more than 1000 in-situ training data points, and most used less than 100. Most of the studies collected their own data, and the resulting data was not generally made public. Collecting in-situ measurements is expensive, and the sample sizes were small. And although the algorithms used were consistent, the framing was not. For example, all of the soil nutrient papers attempted to correlate different properties. All of this hinders the creation of robust deep learning models.

\subsubsection{Recommendations}

An effort to gather existing data and harmonise data collection methods would help to create datasets at a sufficient size to train more robust deep learning models. Additionally, innovations in data collection techniques could also greatly improve data availability as well. For example, refining the soil probes to be produced in quantity, streamlining reading probe outputs and enlisting the help of non-experts.

\subsection{Plant physiology}
\label{section:physiology}

\begin{figure*}
    \centering
    \includegraphics[width=0.7\textwidth]{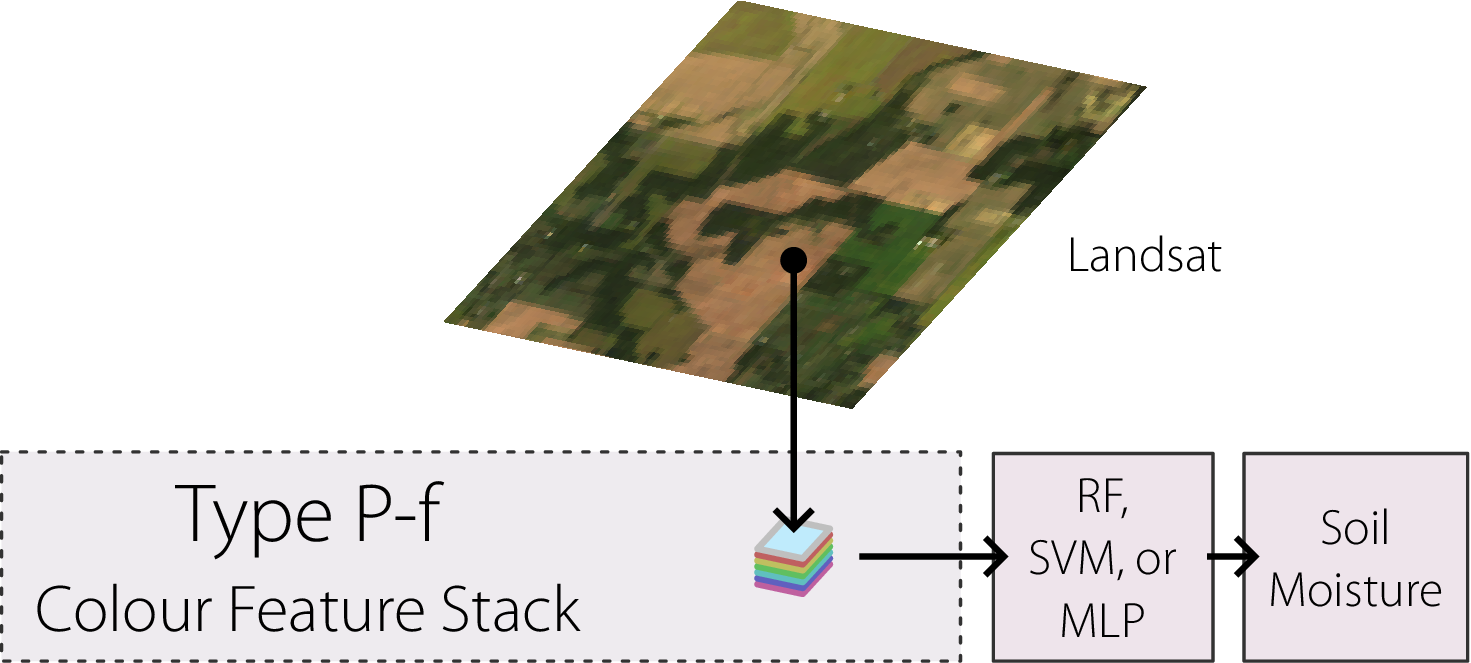}
    \caption{The most common input method for soil health, plant physiology and crop damage detection is to operate on each pixel independently, and treat the pixel as a unordered collection of features (type P-f). This simple method can easily be applied to any pixel-level data, but lacks any contextual information and is used in traditional machine learning methods like RF and SVM, as well as simple MLPs.}
    \label{fig:type-pf}
\end{figure*}

\begin{table*}
    \footnotesize
    \caption{Counts of studies on Plant Physiology. n is the total number of studies. VI counts studies that used vegetation indices. In the labels column: FS = Field survey, IS = Image Survey, MB = Model-based, MBt/FSe = Model-based for train and Field Survey for evaluation. Note that studies may use multiple Models and types in the same study.}
    \centering
    \renewcommand{\arraystretch}{1.2}
    \newcommand{\rot}{\rotatebox{90}}

\begin{tabular}{rc|c|cccc|ccccc|ccccc}
& \textbf{n} & \textbf{VI} & \multicolumn{4}{c|}{\textbf{Labels}} & \multicolumn{5}{c|}{\textbf{Model}} & \multicolumn{5}{c}{\textbf{Type}} \\
& & & \rot{FS} & \rot{MB} & \rot{MBt/FSe} & \rot{Govt.} & \rot{Tree} & \rot{SVM} & \rot{MLP} & \rot{CNN} & \rot{RNN} & \rot{P-f} & \rot{T-f} & \rot{S-i} & \rot{ST-c} & \rot{S-o(P-f)} \\
\hline
\textbf{Canopy Cover/LAI} & 14 & 3 & 6 & 1 & 7 & 0 & 1 & 1 & 14 & 0 & 0 & 10 & 0 & 0 & 0 & 4 \\
\textbf{Other}            &  5 & 4 & 1 & 1 & 1 & 2 & 1 & 1 &  4 & 2 & 1 &  3 & 1 & 2 & 1 & 0 \\
\end{tabular} 

    \label{tbl:agg-canopy-cover}
\end{table*}

Canopy Cover and Leaf Area Index (LAI) are the most commonly predicted plant physiology measurements from satellite imagery using deep learning (14/19; see Table \ref{tbl:agg-canopy-cover}). These two measures are used a proxy for overall plant growth \cite{wu_comparison_2015}. Canopy cover is simply a measurement of the proportion of light being intercepted by the plant, and thus is fairly straight forward to measure with remote sensing. LAI is half of the green leaf area per unit ground surface area \cite{chen_defining_1992}, which incorporates the extent of vertical/overlapping canopy cover. At ground level, LAI is typically measured with a device such as the LAI-2000. This device measures the canopy cover at multiple angles, and uses the assumption of uniformly randomly distributed leaves to calculate LAI. Canopy cover and LAI are highly correlated metrics \cite{nielsen_canopy_2012}, so these are considered together for this review.

\subsubsection{Existing Approaches}

The majority of plant physiology studies reviewed used only MLPs as their primary model without comparing against other machine learning methods (17/19). Out of these, 11/19 used single pixel data (type P-f), 3/19 used aggregated spatial data (type S-o(P-f)) and 1/19 used temporal data as a feature stack (type T-f) (see Figure \ref{fig:type-pf}). There were no examples of modern deep learning for canopy cover or LAI prediction.

There were five plant physiology studies which were not predicting canopy cover or LAI. Instead, they predicted a variety of other properties: water use efficiency \cite{wagle_parameterizing_2016}, gross primary product \cite{wolanin_estimating_2019}, solar-induced flourescence \cite{kira_extraction_2020} and classifying growth stage \cite{thorp_deep_2021, zhao_spatial-aware_2022}. The first three of these works also used MLPs in a similar way. \citet{thorp_deep_2021} was the only work which compared modern deep learning approaches to traditional machine learning.

\subsubsection{Datasets}

The works which did not predict Canopy Cover or LAI had a variety of data sources; mostly government organisations. For example, \citet{thorp_deep_2021} use a large plant growth stage dataset collected by various people using a phone application for Statistics Indonesia. This highlights that larger datasets can be collected when supported by data collection and collation efforts. In contrast, all but one LAI study gathered their data independently. \citet{sun_leaf_2021} used a public dataset of in-situ LAI measurements from the ImagineS field campaign as evaluation data.

\subsubsection{Radiative Transfer models}

To supplement their small datasets, many works predicting LAI used a radiative transfer model (RTM) to generate training data. An RTM is a method to estimate the reflectance from physical properties of the object being imaged. Most of the studies used the PROSAIL model \cite{jacquemoud_prospectsail_2009}, to model the plants and the soil beneath them. It takes as input: the chlorophyll content, the water/dry matter content, the Leaf Area Index, average leaf inclination, ratio of diffuse to direct incident radiation, soil brightness, solar zenith angle, sensor zenith angle and relative azimuth angle; and outputs reflectance. Since the physiological properties are the input to the PROSAIL model, the process of determining one of the physiological properties (e.g. LAI) from the reflectance is called ``model inversion'' in these contexts.

Three quarters of the LAI studies (8/13) collected ground-level physical properties and used an RTM in forward mode to generate training pairs of LAI/reflectance for their MLPs, and then tested them on real data. However, \citet{tomicek_prototyping_2021} found that the PROSAIL RTM predicted reflectances only correlated with Sentinel-2 images between $R^2>0.6$ for red, green and blue bands and up to $R^2>0.9$ for red-edge and NIR bands, which implies fundamental limits on the accuracy of LAI estimation using this method. Despite the acknowledged errors, these studies treat the PROSAIL model as inherently correct, and thus no study compared PROSAIL in reverse mode to a machine learning model on a held-out test set of field survey data.

\subsubsection{Recommendations}

If it is possible to predict LAI from satellite images, it will require more data than an individual author can collect. There is a small public dataset of in-situ LAI measurements\footnote{http://fp7-imagines.eu/pages/services-and-products/ground-data.php} which should be considered in future LAI prediction work, but more such efforts are needed. Data sharing would be another good step for collecting larger datasets for robust training. Of the studies identified, 13 predicted LAI, 12 of which collected their own measurements and did not provide their data publicly.

\subsection{Crop Damage}
\label{section:damage}

\begin{table*}
    \footnotesize
    \caption{Counts of studies on crop damage. VI counts studies that used vegetation indices. In the labels column: FS = Field survey, IS = Image Survey. Note that studies may use multiple Models and types in the same study.}
    \centering
    \renewcommand{\arraystretch}{1.2}
    \newcommand{\rot}{\rotatebox{90}}

\begin{tabular}{rc|c|cc|ccccc|ccccc}
& \textbf{n} & \textbf{VI} & \multicolumn{2}{c|}{\textbf{Labels}} & \multicolumn{5}{c|}{\textbf{Model}} & \multicolumn{5}{c}{\textbf{Type}} \\
& & & FS & IS & \rot{Tree} & \rot{SVM} & \rot{MLP} & \rot{CNN} & \rot{RNN} & \rot{P-f} & \rot{S-o(P-f)} & \rot{T-f} & \rot{T-s} & \rot{S-i} \\
\hline
\textbf{Disease} & 6 & 5 & 6 & 0 & 2 & 2 & 6 & 0 & 0 & 3 & 2 & 1 & 1 & 0 \\
\textbf{Other}   & 4 & 0 & 3 & 2 & 1 & 0 & 2 & 2 & 1 & 1 & 1 & 0 & 0 & 2 \\
\end{tabular} 
    \label{tbl:agg-crop-damage}
\end{table*}

In the reviewed studies, there were two types of crop damage identified: disease/pest damage (6) and physical/generic damage (3). Plants affected by diseases and pests will generally not disappear or die all at once (with notable exceptions \cite{bi_gated_2020}). Instead, the leaves will progressively succumb to damage as the fungus or insect causing the problems grows out of control. Once the changes become wide-spread enough, they become visible via remote sensing \cite{ye_resnet-locust-bn_2020}. Aside from diseases, pests, lack of nutrients and poisoning, plants can also be threatened by extreme weather events. So, some works analyse the physical damage caused by macro-scale physical damage such as large storms \cite{rodriguez_robust_2021} or too many dusty days blocking the sunlight \cite{boroughani_assessment_2022}.

\subsubsection{Existing Approaches}

There were 9 works using deep learning to detect crop damage using satellite data (see Table \ref{tbl:agg-crop-damage}, and of these, only two used CNNs (type S-i) \cite{virnodkar_denseresunet_2021, rodriguez_robust_2021} and one used a GRU (type T-s) \cite{bi_gated_2020}. All such works found improvements using modern deep learning algorithms.

Almost all of the reviewed studies posed the damage caused by disease and physical damage as a classification/segmentation problem (8/9). In these cases, the models were segmenting between "damage" and "no damage". The exception was \citet{rodriguez_robust_2021}, who used a modified ResNet-18 \cite{he_identity_2016} CNN model to detect damage to Coconut crops in the Philippines by comparing tree density estimates before and after Typhoon Goni, making it the only study to measure the extent of the damage per unit area.

The threshold for considering an area to be infected with a disease varied significantly between studies. For example, \citet{yuan_damage_2014} required 80\% of the plants to have visible pustules on the top of the canopy to be considered "infected" with powdery mildew, while \citet{ma_integrating_2019} required more than 10\% of the leafs to be infected to be considered "infected". Obviously a more extensive spread of a disease makes the disease more visible, and hence, predictable. For example, \citet{pignatti_sino-eu_2021} explicitly tested their MLP model at different times after initial infection and noted significantly improved accuracy for later predictions.

\subsubsection{Datasets}

All of the reviewed works collected their data themselves, mostly from field surveys. There were no external datasets mentioned. These studies were all straight-forward applications of well-known deep learning algorithms (mostly MLPs on single-pixel data). All of the studies used medium resolution images (between 1.5m and 30m resolution) to match the resolution of the fields in the field surveys. 

\subsubsection{Recommendations}

Given the progressive nature of plant diseases, it is odd that authors typically chose to frame the problem as a segmentation problem, rather than a regression problem. Under a regression framing, it's possible to use time series analysis to consider temporal relationships and provide a more comprehensive view of the damage. There is no good evidence for which temporal model to use for such temporal data; there was one work that found a GRU improved over MLP or RF models \cite{bi_gated_2020}. Authors could try LSTMs and transformers as well.

\subsection{Yield}
\label{section:yield}

The task of yield prediction is to predict the average yield per hectare for a parcel of land. Unlike LULC, which can be identified by a human expert from a satellite image moderately reliably \cite{mcnairn_integration_2009}, yield is generally measured at either the county-level by government bodies, or at the field-level by individual researchers. For the purposes of this review, "county" refers to anything larger than a field (e.g. ``municipality'', ``province'', ``state'', etc.) due to their varying meaning across different studies/governments. 

\begin{figure*}
    \centering
    \includegraphics[width=\textwidth]{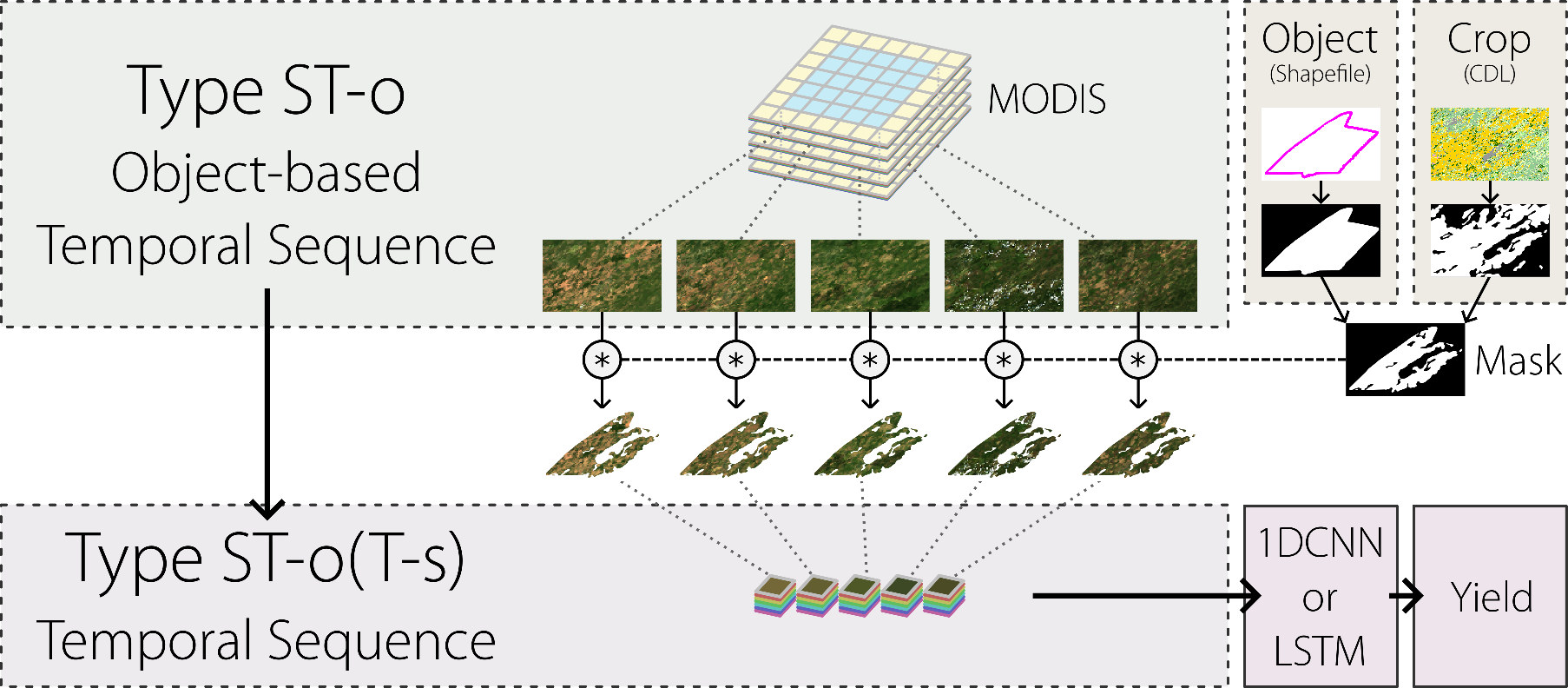}
    \caption{The most common input method for county-level yield prediction is to use object boundaries to select and aggregate pixels from spatiotemporal data (type ST-o), and reinterpret this data as a simple temporal sequence of county-averaged ``pixels'' (type T-s). Data of this form is denoted type ST-o(T-s). This T-s data can then be used in models like 1DCNNs or LSTMs. Many works will mask the image data by both the shapefile and crop type using the USDA's CDL \cite{usda_national_2022} or equivalent, but other works indiscriminately aggregate over all pixels within the county boundary, including pixels from different crop types and some non-farmland. Rectangles with dotted borders denote different data, and rectangles with solid borders denote algorithms and their outputs.}
    \label{fig:type-stots}
\end{figure*}

\begin{table*}
    \footnotesize
    \caption{Counts of studies on Yield prediction. VI counts studies that used vegetation indices. Clim. counts studies that used climate variablesIn the labels column: FS = Field survey, and Govt. = Government. Note that studies may use multiple Models and types in the same study. Only types used in more than one study are shown.}
    \centering
    \renewcommand{\arraystretch}{1.2}
    \newcommand{\rot}{\rotatebox{90}}

\begin{tabular}{rc|cc|ccc|ccccc|ccccccccc}
& \textbf{n} & \rot{\textbf{VI}} & \rot{\textbf{Clim.}} & \multicolumn{3}{c|}{\textbf{Labels}} & \multicolumn{5}{c|}{\textbf{Model}} & \multicolumn{9}{c}{\textbf{Type}} \\
& & & & \rot{FS} & \rot{Govt.} & \rot{Other} & \rot{Tree} & \rot{SVM} & \rot{MLP} & \rot{CNN} & \rot{RNN} & \rot{P-f} & \rot{T-f} & \rot{ST-c} & \rot{S-o(P-f)} & \rot{ST-o(P-f)} & \rot{ST-o(T-f)} & \rot{ST-o(T-s)} & \rot{ST-o(T-i)} & \rot{ST-o(T-c)} \\
\hline
\textbf{County-level} & 23 & 21 & 20 & 1 & 22 & 3 & 16 & 6 & 11 & 7 & 12 & 0 & 0 & 0 & 0 & 1 & 18 & 14 & 2 & 2 \\
\textbf{Field-level}  &  7 &  7 &  2 & 6 &  0 & 1 &  3 & 2 &  5 & 1 &  2 & 0 & 0 & 0 & 2 & 3 &  2 &  2 & 0 & 0 \\
\textbf{Plot-level}   &  2 &  1 &  1 & 2 &  0 & 0 &  1 & 1 &  2 & 1 &  0 & 0 & 1 & 1 & 0 & 1 &  0 &  0 & 0 & 0 \\
\textbf{Pixel-level}  &  2 &  2 &  1 & 2 &  0 & 1 &  0 & 0 &  1 & 1 &  1 & 1 & 0 & 0 & 0 & 0 &  0 &  0 & 0 & 0 \\
\textbf{Other}        &  7 &  5 &  3 & 5 &  1 & 1 &  1 & 1 &  7 & 1 &  1 & 3 & 2 & 1 & 0 & 0 &  0 &  0 & 0 & 1 \\
\end{tabular} 
    \label{tbl:agg-yield}
\end{table*}

\subsubsection{Existing approaches}


Over half of the reviewed yield prediction studies were at the county-level (23/41), and almost all of these used MODIS (21/23). The other two county-level yield prediction studies used Landsat/Sentinel images (see Table \ref{tbl:agg-yield}). There were 4 studies that measured grassland biomass, and one that merely measured county-wide NDVI. The rest (13/41) investigated yield at a farm-level or smaller, and generally used Landsat or Sentinel images (10/13), and some used WorldView-3 images (2/13).

County-level yield prediction works universally used ST-o data and aggregated to T-f, T-s or T-i data (see Figure \ref{fig:type-stots}). This removes all spatial information and poses it as a relationship between a temporal sequence of spatially-aggregated colour information (NDVI or reflectances) and the final yield. The most common aggregation method was to take the average, but a naive average includes significant non-crop information from bodies of water, residential areas, and other unfarmed land. So, some authors used LULC products (e.g. CDL) to exclude MODIS pixels which did not contain the crops they were interested in \cite{li_estimating_2007}. Even when using such a method, there are still significant sources of error in the aggregation: the fields may be mislabelled, and fields that are double-cropped may or may not contain the crop of interest at the time the image is taken.

Field-level yield prediction also used ST-o(T-s) data in much the same manner as for county-level. But, instead of using MODIS pixels which would be too large to describe individual fields, they used Landsat/Sentinel images. And to avoid including non-crop information they generally only include pixels wholly within the field boundaries to avoid ``border'' effects. 

Aggregating with a naive average loses a lot of information, so a few authors investigated using alternative methods. Instead of reducing each band to a single value, \citet{khaki_simultaneous_2021} created a histogram for each colour band and then used a 2DCNN on each county's histogram. At smaller scales another option is to simply not do any aggregation at all. \citet{engen_farm-scale_2021} set all pixels outside the field of interest to zero, used a 2DCNN on each image directly, and a GRU to process across images.

The county-level works used RFs, SVMs, MLPs, 1DCNNs, LSTMs and even some 2DCNNs. They had many comparisons, but there were no obvious trends. This is unlike other tasks where modern deep learning algorithms showed almost universal improvements. More specifically, 10 studies compared LSTMs to traditional machine learning models and the LSTM performed better in only 6 of them, with the other four finding SVMs and trees to perform better. Similarly, in field-level yield prediction, only one of the two studies comparing LSTMs found improved performance over traditional machine learning.

\subsubsection{Climatic and soil variables}
\label{section:yield-soil}

Many county-level studies noted that using only reflectance wasn't sufficient to model yield. So, nearly all of the county-level studies used climatic variables (e.g. ground temperature and precipitation) and soil properties (e.g. soil particle size, soil moisture content) in addition to the reflectances/VIs (22/23), and universally noted improved performance when compared to not using these variables \cite{cai_integrating_2019, wang_combining_2020, wolanin_estimating_2020}. However, there was no consistency in which soil variables were used, nor how the temperature was encoded, nor did any works compare their climatic data encoding to other encodings. For example, the temperature was variously classified/aggregated into Growing Degree Days, Killing Degree Days and Freezing Degree Days before being fed into the model (e.g. \cite{zhang_integrating_2021}). It was also common to use various different temperature statistics (e.g. min/mean/max) and some studies even combined multiple sources of meteorological data (e.g \cite{ju_optimal_2021}). Thus no particular climatic or soil properties are recommended here.

In contrast, few smaller-scale works used climatic/soil variables (2/7). This is because most operated in a single location, and such weather/soil information would be identical for all examples in their dataset. When operating at smaller scales but also across multiple locations climatic variables can be used. In such cases, it is better to interpolate weather data between stations, instead of selecting the values from the nearest station \cite{engen_farm-scale_2021}.

\subsubsection{Datasets}

Near universally, the county-level works used government-published data (22/23). Indeed, the county-level delineation is likely so popular precisely because that is the scale that governments collect aggregated yield information. Several of the county-level studies have over 1000 counties recorded for over 10 years. This scale of data is easy to obtain for a researcher, as it is all publicly downloadable.

In contrast, the reviewed field-level studies used small datasets containing less than 100 datapoints collected from field studies conducted specifically for that work. There were two notable exceptions. \citet{zhang_integrating_2021} used a particularly large field survey containing 11000 data points. \citet{engen_farm-scale_2021} predicted at farm-scale across many thousand farms, using data collected by the Norwegian government from the farmers directly, similar to the other LULC sources from European governments.

At the finest resolution, two works utilised advanced combine harvesters that could measure yield as it was collected through the field. This allowed them to create dense spatial maps of yield and correlate this with the satellite images. By collecting a dense map, more information is extracted from each field, aggregation within a field is not necessary, and CNNs become an obvious choice.

\subsubsection{Analysis}

It is not clear why modern deep learning did not consistently outperform traditional machine learning for yield prediction, like it did for other tasks. The datasets are generally larger than those used in other tasks (not including LULC). But, most work in yield prediction operates on highly aggregated information. And \citet{wolanin_estimating_2020} found that using the average yield for a county as the prediction performs surprisingly well. So, perhaps performance using averaged reflectance is effectively saturated and whether one model does better than another is somewhat random.

The ability to describe a complex trait like yield with averaged reflectance values obviously has limits. So, instead of averaging to a single value per colour, \citet{khaki_simultaneous_2021} collected a histogram for each county, which improved results and showed that there is potential to explore novel aggregation techniques. 

Farm/field-level yield prediction is still relatively new, and requires more support to collate large enough datasets to train robust deep learning models. And when such models become reliable, county-level yield prediction will naturally follow, albeit with significantly more computation required.

\subsubsection{Recommendations}

More investigation is needed to explain why modern deep learning methods have been roughly on par with traditional machine learning methods. The majority of reviewed studies compared only machine learning methods; it should be standard practice to include simple non-machine learning baselines to contextualise the results and show the benefit of machine learning.

Additionally, moving beyond modelling reflectance as a single averaged value is fertile ground to improve yield prediction. In particular, exploring ways to construct the problem to allow deep learning models to identify more complex relationships.

\section{Discussion}

This systematic review covers 193 studies at the intersection of Deep Learning on Satellite Imagery for Agriculture. These were identified using Clarivate's Web of Science and the search criteria outlined in Section \ref{section:systematic-search}. As such it will not cover all possible works at this intersection, and all claims about numbers of studies refer only to those identified by the search methodology.

In this review, we have observed that modern deep learning methods have become dramatically more popular in the last few years. This trend is likely to continue, as the research appears to be following generic computer vision. However, generic computer vision research has been partially propelled by large public benchmark datasets, which are mostly lacking in the agricultural tasks identified in this review, especially monitoring soil health, plant physiology and crop damage. Without accessible large datasets, methods are not easily comparable and deep learning in these areas is constrained to those who have the resources to conduct large field surveys.

\subsection{Datasets}

There has never been more data available for training deep learning on satellite images, but this availability is very uneven, and underutilised. So there are two issues. First, there is a lack of labelled data for most agricultural tasks. And second, even when there is an abundance of labels, authors prefer to create their own dataset rather than compare results on existing datasets. Crop segmentation has several benchmark datasets (see Table \ref{tbl:datasets}; Section \ref{section:LULC}), but they have generally not been adopted. Only 3 out of the 193 studies compared their models against other published models on benchmark datasets. But, of these three, none compared against another study on the same dataset. Benchmarks are critically important for effective model comparison, and consequently driving algorithm innovation. But these benefits can only be realised if studies engage with the datasets as a competition, and diligently compare against existing algorithms.

All of the crop segmentation datasets identified were semantic segmentation. All semantic segmentation datasets with shapefile labels could become a panoptic segmentation dataset by additionally identifying individual fields. But, after inspecting each named dataset -- and each reviewed study using those datasets -- we note that none of them did so. Similarly, a multi-year segmentation dataset could become a change detection dataset, and similarly, no reviewed study did so.

All other tasks did not have benchmark datasets, but some had continuously updated data sources. There are some sources of public data for Leaf Area Index (LAI), soil moisture and county-level yield. But no researchers have published a specific list of examples to create a benchmark dataset. But, where do these large datasets come from? Large datasets for agricultural tasks require a significant investment in data collection, and a bit of ingenuity. In the reviewed studies, the large datasets were invariably the result of a dedicated governmental effort plus some way to distribute the work. For example, for growth stage classification in rice paddies, \citet{thorp_deep_2021} collected data from across the whole of Indonesia, as reported by 4778 crop surveyors and collated efficiently through a phone application. And, for crop segmentation, governments have incentivised (or compelled) farmers to self-report their crops. While that level of support/adoption might be out of reach for most researchers, there's an unaddressed gap. Innovations in data collection could make medium sized datasets feasible, and are a necessary prerequisite to begin large-scale data collection besides.

For the Soil health (21/25), Plant physiology (15/19), Crop damage (8/9) and sub-county-level Yield prediction tasks (14/19), the authors created their own datasets by field surveys. For these tasks, there were no examples of authors using datasets collected by other authors for the same task. The rest were instead already collected by the government for another purpose (8/72), or automatically generated via a different algorithm (5/72). Although individually small collections, many of these data collection efforts could be released by their authors and included in future works, or could be used for guidance of data collection processes.

\subsection{Models}


There were 53 studies identified across soil health, plant physiology and crop damage tasks, and among these only eight of them used modern deep learning algorithms and most of those were after 2020. Of these eight, only four compared the modern deep learning algorithm to a tree/forest, SVM or MLP, but in all four cases the modern deep learning algorithm performed better. For soil health, plant physiology and crop damage tasks, the most common method was relatively straightforward use of P-f models. Since the search criteria specifically selected for deep learning algorithms, almost all of the reviewed papers for these tasks used MLPs (48/53). Very few studies of these less popular tasks used spatial algorithms (6/53), and almost none used time series algorithms (3/53), despite several using freely available satellite imagery with revisit times on the order of days.

The vast majority of modern deep learning methods were for LULC tasks, in which there was a very noticeable trend of modern deep learning algorithms performing better. In crop segmentation, modern deep learning algorithms were compared with trees, SVMs or MLPs in 25 studies, and performed better in all but one case. Vision Transformers \cite{dosovitskiy_image_2021} and variants were only found to be used in crop segmentation. Surprisingly, although these have performed very well in generic computer vision tasks, they did not show any particular improvement over CNNs and RNNs. Transformers have been noted to require much more data than CNNs to achieve state-of-the-art performance, and it's possible that the transformers used in the reviewed studies have not been trained with enough data. So, there is room for further investigation of this phenomenon. Given the large volume of unlabelled data, perhaps utilising unsupervised or semi-supervised training methods could bridge the gap in data scale.

For as many examples of straightforward uses of standardised forms of modern deep learning algorithms (e.g. UNet fine-tuned on type S-i data \cite{pedrayes_evaluation_2021}), there were also many examples of specific ideas being directly transferred and adapted from generic computer vision to the satellite setting (e.g. discriminator loss \cite{li_adversarial_2021}, neural ordinary differential equations \cite{metzger_crop_2022}, channel attention maps \cite{tang_channel_2022}). However, there was only one main variation observed that explicitly took advantage of the differences between satellite imagery and ground-based imagery: several works utilised the higher spectral resolution and convolved over the spectral dimension (types P-s, T-i and S-c; e.g. \cite{kussul_deep_2017, debella-gilo_mapping_2021, saralioglu_semantic_2022}). There is a gap here; satellite imagery is not the same as ground-based images or video. There is temporal continuity at every pixel. There are spatial patterns at hugely varied scales. There is an order of magnitude more data available, and it is all associated with physical locations. All of these properties are not being taken advantage of in existing work tackling agricultural tasks, and are interesting directions for future research.



\section*{Acknowledgements}
This work has been supported by the SmartSat CRC, whose activities are funded by the Australian Government’s CRC Program.

Certain data included herein are derived from Clarivate Web of Science. © Copyright Clarivate 2022. All rights reserved.

\bibliography{references}{}
\bibliographystyle{plainnat}

\begin{IEEEbiography}[{\includegraphics[height=1.25in]{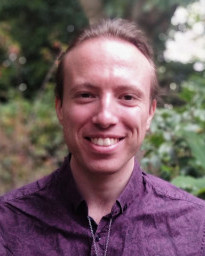}}]{Brandon Victor}
Brandon is a PhD student working under Dr. Aiden Nibali and Dr. Zhen He at La Trobe University. He has experience with every stage in the development and deployment of deep learning solutions for custom projects. And, he has worked on a variety of computer vision and deep learning projects, such as sports analytics, and now, satellites and agriculture.
\end{IEEEbiography}

\begin{IEEEbiography}[{\includegraphics[height=1.25in]{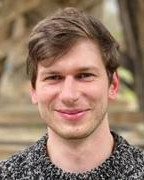}}]{Dr. Aiden Nibali}
Dr Nibali's core area of research is the development of computer vision systems utilising artificial intelligence. He has many years of practical programming experience and deep neural network development. His work on AI projects is often cross-disciplinary, incorporating other research areas such as medical imaging, alcohol research, agriculture, and sports analysis.
\end{IEEEbiography}

\begin{IEEEbiography}[{\includegraphics[height=1.25in]{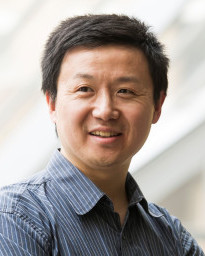}}]{Dr. Zhen He}
Dr He is an Associate Professor of computer science at La Trobe University. He leads a deep learning research group focused on computer vision tasks. Dr He’s team applies deep learning to a number of different application areas including: medical image analysis; precision agriculture; sports analytics; and alcohol exposure detection. His current primary focus is on applying AI to analyze digital pathology images.
\end{IEEEbiography}

\end{document}


\section{Supplementary}

\subsection{Search terms}

The full list of crop names searched for: Wheat, Corn, Maize, Orchard, Coffee, Vineyard, Soy, Rice, Cotton, Sorghum, Peanut*, Tobacco, Barley, Grain, Rye, Oat, Millet, Speltz, Canola, Flaxseed, Mustard, Alfalfa, Camelina, Beans, Potato*, Sugar*, Vegetable*, Fruit, Onion*, Cucumber*, Peas, Lentils, Tomato*, Hops, Herbs, Peach*, Apple*, Grape*, Citrus, Pecan*, Almond*, Walnut*, Pear*, Aquaculture, Perrenial, Pistachio*, Triticale, Carrot*, Asparagus, Garlic, Cantaloupes, Prune*, Orange*, Olive*, Broccoli, Avocado*, Pepper*, Pomegranate*, Nectarine*, Plum*, Squash, Apricot*, Vetch, Lettuce, Pumpkin, Cabbage, Celery, Radish*, Turnip*, Eggplant*, Gourd*

\subsection{Dataset links}
\begin{table*}[h]
\small
\centering
\caption{Identified crop segmentation satellite benchmark datasets and data sources, along with links to download the dataset.}
\newcommand{\rot}{\rotatebox{90}}

\begin{tabular}{p{150pt}|p{350pt}}
\textbf{Name} & \textbf{Link} \\
\hline
LUCAS \citep{dandrimont_lucas_2021} & https://data.jrc.ec.europa.eu/dataset/cfe66a0c-bdee-4074-96e1-a2f7030b9515 \\
\hline
Munich dataset \citep{ruswurm_multi-temporal_2018} & https://github.com/MarcCoru/MTLCC \\
\hline
BreizhCrops \citep{breizhcrops2020} & https://github.com/dl4sits/BreizhCrops \\
\hline
ZueriCrop \citep{turkoglu_crop_2021} & https://polybox.ethz.ch/index.php/s/uXfdr2AcXE3QNB6 \\
\hline
DENETHOR \citep{kondmann_denethor_2021} & https://github.com/lukaskondmann/DENETHOR \\
\hline
Reunion Island \citep{dupuy_reunion_2020} & https://dataverse.cirad.fr/dataset.xhtml?persistentId=doi:10.18167/DVN1/YZJQ7Q \\
\hline
Campo Verde \citep{sanches_campo_2017} & https://ieee-dataport.org/documents/campo-verde-database \\
\hline
UOS2 \citep{pedrayes_evaluation_2021} & https://zenodo.org/record/4648002 \\
\hline
PASTIS-R \citep{garnot_multi-modal_2022} & https://zenodo.org/record/5735646 \\  
\hline
Sen4AgriNet \citep{sykas_sentinel-2_2022} & https://github.com/Orion-AI-Lab/S4A \\ 
\hline
French Land Parcel Identification System & https://www.data.gouv.fr/fr/datasets/registre-parcellaire-graphique-rpg-contours-des-parcelles-et-ilots-culturaux-et-leur-groupe-de-cultures-majoritaire/ \\
\hline
Cropland Data Layer \cite{usda_national_2022} & https://nassgeodata.gmu.edu/CropScape/  \\
\hline
Corine Land Cover & https://land.copernicus.eu/pan-european/corine-land-cover \\
\hline
WorldCereal & https://zenodo.org/records/7593734 \\
\hline

\end{tabular}
\label{tbl:datasets-links}
\end{table*}

Each dataset from the LULC section can be downloaded from the links in Table \ref{tbl:datasets-links}.

\subsection{Tagged database}

All references for this review are made available as a Zotero rdf file. It includes notes and tags on all studies.

\pagebreak

\subsection{All Crop Segmentation studies}

Here, we list all of the crop segmentation studies.

Studies on crop segmentation (up to 2020). N is the number of classes that the study distinguished between (number of agricultural classes in brackets). VI stands for vegetative indices, and indicates whether VIs were used (possibly in combination with other features). GSD = Ground Spatial Distance/resolution. In the labels column, a $\dagger$ indicates that the labels are from a dataset (see Table \ref{tbl:datasets-links}). The best performing model type in each study is underlined. Where no model is underlined, no one model performed best. Where two model types are underlined, it means the model had components of both types. In the Tree column: RF = Random Forest, and a tick means any other kind of tree. In the CNN column: nD = nDCNN, including non-spatial CNNs. In the RNN column: L = LSTM, G = GRU, and nD = nD ConvRNN (ConvGRU or ConvLSTM). In the Attn column: T = Transformer, P = PSE-TAE \cite{garnot_satellite_2020} and a tick indicates any other transformer-like attention.
{
\small
\centering
\newcommand{\rot}{\rotatebox{90}}

\begin{tabular}{%
    >{\raggedleft}p{80pt}|
    >{\raggedleft}p{28pt}|
    >{\raggedright}p{50pt}|
    c|
    p{23pt}|
    >{\raggedright}p{60pt}|
    >{\raggedright\centering}p{12pt}|
    c|
    c|
    >{\raggedright\centering}p{12pt}|
    >{\raggedright\centering}p{12pt}|
    c|
    p{23pt}
}

&  &  &  &  &  & \multicolumn{6}{c|}{\textbf{Model}} & \\
\textbf{Study} & \textbf{N} & \textbf{Images} & \textbf{VI} & \textbf{GSD} & \textbf{Labels} & \rot{Tree} & \rot{SVM} & \rot{MLP} & \rot{CNN} & \rot{RNN} & \rot{Attn} & \textbf{Type} \\
\hline

\cite{ban_synergy_2003} & 8 (8) & ERS-1, Landsat &  & 30m & Field Survey &  &  & \underline{\checkmark} &  &  &  & ST-o \newline (T-f) \\
\rowcolor{gray!10} \cite{karkee_quantifying_2009} & 2 (1) & MODIS & \checkmark & 1km & Model-based &  &  & \underline{\checkmark} &  &  &  & P-f \\
\cite{pena_object-based_2014} & 9 (9) & ASTER & \checkmark & 15m & Priv. govt. & \checkmark & \underline{\checkmark} & \checkmark &  &  &  & S-o \newline (P-f) \\
\rowcolor{gray!10} \cite{kumar_comparison_2015} & 13 (7) & LISS IV &  & 5.8m & Field Survey &  & \underline{\checkmark} & \checkmark &  &  &  & P-f \\
\cite{loew_decision_2015} & 10 (7) & RapidEye & \checkmark & 6.5m & Field Survey & RF \checkmark & \underline{\checkmark} & \checkmark &  &  &  & ?-f \\
\rowcolor{gray!10} \cite{kussul_deep_2017} & 11 (7) & Sentinel-1, Landsat-8 &  & 30m & Field Survey & RF &  & \checkmark & 1D \underline{2D} &  &  & P-f \newline P-s \newline ST-i \\
\cite{shelestov_exploring_2017} & 13 (8) & Landsat-8 &  & 30m & Field Survey & RF & \checkmark & \underline{\checkmark} &  &  &  & S-o \newline (P-f) \\
\rowcolor{gray!10} \cite{benedetti_m3fusion_2018} & 13 (4) & Sentinel-2, SPOT6/7 & \checkmark & 20m, 2m & Reunion Island$\dagger$ & RF &  &  & \underline{2D} & \underline{G} &  & ST-f \newline ST-c \\
\cite{ji_3d_2018} & 9 (4) & GaoFen-1/2 & \checkmark & 15m, 4m & Image Survey &  & \checkmark &  & 2D \underline{3D} &  &  & S-i \newline ST-c \\
\rowcolor{gray!10} \cite{kussul_crop_2018} & 13 (8) & Sentinel-1, Landsat-8 &  & 10m, 30m & Field Survey &  &  & \underline{\checkmark} &  &  &  & P-f \\
\cite{ndikumana_deep_2018} & 11 (8) & Sentinel-1 &  & 20m & Image Survey & RF & \checkmark &  &  & L \underline{G} &  & ST-o \newline (T-f \newline T-s) \\
\rowcolor{gray!10} \cite{interdonato_duplo_2019} & 13 (4) & Sentinel-1/2 & \checkmark & 10m & Reunion Island$\dagger$, Koumia$\dagger$ & \underline{RF} &  &  & \underline{2D} & L \underline{G} 2D &  & T-f \newline T-s \newline ST-c \\
\cite{sidike_dpen_2019} & 19 (11) & WV-3 &  & 0.3m & Field Survey & RF & \checkmark &  & \underline{1D} &  &  & P-f \newline P-s \\
\rowcolor{gray!10} \cite{teimouri_novel_2019} & 16 (13) & Sentinel-1 &  & 10m & Model-based &  &  &  & \underline{2D} & \underline{2D} &  & ST-c \\
\cite{xie_deep_2019} & 15 (12) & GaoFen-1 &  & 2m & Image Survey & RF &  &  & \underline{2D} &  &  & S-i \\
\rowcolor{gray!10} \cite{zhao_evaluation_2019} & 5 (5) & Sentinel-1 &  & 10m & Field Survey & RF &  &  & \underline{1D} & L G &  & T-f \newline T-s \\
\cite{zhou_long-short-term-memory-based_2019} & 7 (7) & Sentinel-1 &  & 10m, 5.1m & Image Survey & RF & \checkmark &  &  & \underline{L} &  & ST-o \newline (T-f \newline T-s) \\
\rowcolor{gray!10} \cite{mazzia_improvement_2020} & 3 (3) & Sentinel-2 & \checkmark & 10m & LUCAS & RF \checkmark & \checkmark &  & \underline{2D} & \underline{L} &  & T-f \newline T-i \\
\cite{nguyen_monitoring_2020} & 2 (1) & Sentinel-2, Landsat-8 &  & 30m, 10m & Field Survey &  &  &  & \underline{2D} & \underline{L} &  & ST-c \\
\rowcolor{gray!10} \cite{ruswurm_self-attention_2020} & 23 (18) & Sentinel-2 &  & 10m & Pub. govt. & RF &  &  & 1D 2D & L G & T & ST-o \newline (T-f \newline T-s) \\
\cite{wang_mapping_2020} & 10 (10) & Sentinel-1/2 & \checkmark & 10m & Field Survey & RF &  &  & \underline{1D} 3D &  &  & T-f \newline T-s \newline ST-c \\
\rowcolor{gray!10} \cite{xu_deepcropmapping_2020} & 3 (2) & Landsat-8 &  & 30m & CDL & RF &  & \checkmark &  & \underline{L} & T & T-f \newline T-s \\

\end{tabular}
}

\pagebreak

Studies on crop segmentation (year 2021).
{
\small
\centering
\newcommand{\rot}{\rotatebox{90}}

\begin{tabular}{%
    >{\raggedleft}p{75pt}|
    >{\raggedleft}p{20pt}|
    >{\raggedright}p{48pt}|
    c|
    p{20pt}|
    >{\raggedright}p{65pt}|
    >{\raggedright\centering}p{12pt}|
    c|
    c|
    >{\raggedright\centering}p{12pt}|
    >{\raggedright\centering}p{12pt}|
    c|
    p{37pt}
}

&  &  &  &  &  & \multicolumn{6}{c|}{\textbf{Model}} & \\
\textbf{Study} & \textbf{N} & \textbf{Images} & \textbf{VI} & \textbf{GSD} & \textbf{Labels} & \rot{Tree} & \rot{SVM} & \rot{MLP} & \rot{CNN} & \rot{RNN} & \rot{Attn} & \textbf{Type} \\
\hline

\cite{chelali_deep-star_2021} & 4 (4) & Sentinel-2 &  & 10m & Pub. govt. &  &  &  & 1D \underline{2D} & L 2D &  & ST-o \newline (ST-i) \\
\rowcolor{gray!10} \cite{choung_comparison_2021} & 3 (2) & KOMPSAT-3 &  & 0.7m & Image Survey &  & \checkmark & \underline{\checkmark} &  &  &  & S-i \\
\cite{debella-gilo_mapping_2021} & 3 (1) & Sentinel-2 &  & 20m & Pub. govt. &  &  & \checkmark & 1D \underline{2D} &  &  & T-f T-s T-i \\
\rowcolor{gray!10} \cite{fernandez-sellers_finding_2021} & 111 (?) & Sentinel-2 &  & 10m & Pub. govt. &  &  & \underline{\checkmark} &  &  &  & T-f \\
\cite{fu_new_2021} & 4 (2) & GaoFen-1 &  & 16m & Field/Image Survey &  &  &  & \underline{2D} &  &  & S-i \\
\rowcolor{gray!10} \cite{gallo_sentinel_2021} & 17 (17) & Sentinel-2 & \checkmark & 10m & Munich$\dagger$ &  &  &  & \underline{3D} &  &  & ST-c \\
\cite{hamer_replacing_2021} & 2 (1) & DMC &  & 32m & Image Survey, Model-based & RF &  &  & \underline{2D} &  &  & P-f \newline S-i \\
\rowcolor{gray!10} \cite{jin_extraction_2021} & 2 (1) & PlanetScope & \checkmark & 3m & Field Survey & \underline{RF} & \checkmark & \checkmark &  &  &  & P-f \\
\cite{laban_sparse_2021} & 8 (6) & Sentinel-2 &  & 10m & Field Survey &  &  &  & \underline{2D} &  &  & P-f (S-i) \\
\rowcolor{gray!10} \cite{lei_docc_2021} & 2 (2) & Zhuhai-1, Sentinel-1/2 &  & 10m & Field/Image Survey &  &  &  & \underline{2D} &  &  & S-i \\
\cite{li_adversarial_2021} & 3 (2) & Landsat-8 &  & 30m & CDL &  & \checkmark &  & \underline{2D} & \underline{L} &  & ST-c \\
\rowcolor{gray!10} \cite{lozano-tello_crop_2021} & 2 (1) & Sentinel-2 &  & 10m & Priv. govt. &  &  & \underline{\checkmark} &  &  &  & P-f \\
\cite{martini_domain-adversarial_2021} & 9 (7) & Sentinel-2 &  & 10m & BreizhCrops$\dagger$ &  &  &  &  &  & \underline{T} & ST-o \newline (T-s) \\
\rowcolor{gray!10} \cite{meng_deep_2021} & 6 (2) & Zhuhai-1, Sentinel-2 & \checkmark & 10m & Field/Image Survey &  &  &  & 1D 2D \underline{3D} &  &  & P-s \newline S-i \newline S-c \\
\cite{moreno-revelo_enhanced_2021} & 11 (6) & Sentinel-1, Landsat-8 &  & 30m & Campo Verde$\dagger$ &  &  &  & \underline{2D} & L &  & S-i S-s \\
\rowcolor{gray!10} \cite{moumni_machine_2021} & 6 (4) & Sentinel-1/2 & \checkmark & 10m & Field/Image Survey & RF & \underline{\checkmark} & \checkmark &  &  &  & T-f \\
\cite{mukharamova_estimating_2021} & 12 (8) & MODIS & \checkmark & 250m & Pub. govt. & RF &  & \checkmark &  & \underline{L} &  & T-s \\
\rowcolor{gray!10} \cite{ofori-ampofo_crop_2021} & 12 (8) & Sentinel-1/2 &  & 10m & Pub. govt. &  &  &  &  &  & \underline{P} & ST-c \\
\cite{pedrayes_evaluation_2021} & 11 (4) & Sentinel-2 &  & 10m & UOS2$\dagger$ & RF & \checkmark &  & \underline{2D} &  &  & P-f \newline S-i \\
\rowcolor{gray!10} \cite{quinton_crop_2021} & 20 (18) & Sentinel-2 &  & 10m & Pub. govt. &  &  &  &  &  & \underline{P} & ST-c \\
\cite{rahimi-ajdadi_remote_2021} & 4 (2) & Landsat-5/7/8 &  & 30m & Field/Image Survey, Model-based &  &  & \underline{\checkmark} &  &  &  & P-f \\
\rowcolor{gray!10} \cite{rawat_deep_2021} & 5 (4) & Sentinel-2, Landsat-8 &  & 10m & Field Survey &  &  &  & \underline{1D} & 1D &  & T-s \\
\cite{thorp_deep_2021} & 6 (3) & Sentinel-1/2 & \checkmark & 10m & Priv. govt. & RF & \checkmark & \checkmark & 2D & \underline{L} G 2D &  & P-f S-i \newline T-s ST-c \\
\rowcolor{gray!10} \cite{turkoglu_crop_2021} & 48 (41) & Sentinel-2 &  & 10m & ZueriCrop$\dagger$ & RF &  &  & 2D & L \underline{2D} & T & ?-f S-i \newline T-s ST-c \\
\cite{wu_winter_2021} & 2 (1) & Sentinel-1 &  & 10m & Model-based &  &  &  & \underline{2D} &  &  & S-i \\
\rowcolor{gray!10} \cite{xie_integration_2021} & 2 (1) & MODIS & \checkmark & 250m & Image Survey & RF &  &  & 1D & \underline{L} &  & T-f \newline T-s \\
\cite{zhang_rapid_2021} & 3 (2) & Sentinel-2, Landsat-8 & \checkmark & 10m, 30m & Pub. govt., Field Survey &  &  & \underline{\checkmark} &  &  &  & T-f \\
\rowcolor{gray!10} \cite{zhao_evaluation_2021} & 7 (6) & Sentinel-2 &  & 10m & Field/Image Survey &  &  &  & \underline{1D} & L G 1D &  & T-s \\

\end{tabular}
}

\pagebreak

Studies on crop segmentation (year 2022).
{
\small
\centering
\newcommand{\rot}{\rotatebox{90}}

\begin{tabular}{%
    >{\raggedleft}p{70pt}|
    >{\raggedleft}p{28pt}|
    >{\raggedright}p{60pt}|
    c|
    p{23pt}|
    >{\raggedright}p{60pt}|
    >{\raggedright\centering}p{12pt}|
    c|
    c|
    >{\raggedright\centering}p{12pt}|
    >{\raggedright\centering}p{12pt}|
    c|
    p{22pt}
}

&  &  &  &  &  & \multicolumn{6}{c|}{\textbf{Model}} & \\
\textbf{Study} & \textbf{N} & \textbf{Images} & \textbf{VI} & \textbf{GSD} & \textbf{Labels} & \rot{Tree} & \rot{SVM} & \rot{MLP} & \rot{CNN} & \rot{RNN} & \rot{Attn} & \textbf{Type} \\
\hline

\cite{asming_processing_2022} & 9 (2) & Landsat-8, Sentinel-2 &  & 30m, 10m & Image Survey &  &  & \underline{\checkmark} &  &  &  & P-f \\
\rowcolor{gray!10} \cite{desloires_positive_2022} & 2 (1) & Sentinel-2 &  & 10m & Pub. govt. &  &  &  &  & \underline{G} &  & T-s \\
\cite{fontanelli_early-season_2022} & 10 (9) & COSMO-SkyMed &  & 15m & Field Survey &  &  &  & 1D \underline{3D} &  &  & T-s \newline ST-c \\
\rowcolor{gray!10} \cite{garnot_multi-modal_2022} & 30 (27) & Sentinel-1/2 &  & 10m & Pub. govt. &  &  &  & \underline{2D} &  & \underline{P} & ST-c \newline ST-o \newline (T-s) \\
\cite{jiang_crop_2022} & 10 (3) & Sentinel-2 &  & 10m & CDL & RF \checkmark & \checkmark & \underline{\checkmark} &  &  &  & P-f \\
\rowcolor{gray!10} \cite{lei_multi-temporal_2022} & 2 (1) & SPOT, Sentinel-1 & \checkmark & 6m, 10m & Pub. govt. &  & \checkmark & \underline{\checkmark} &  &  &  & ST-f \\
\cite{li_crop_2022} & 3 (2) & RadarSat-2 & \checkmark & 8m & Field Survey &  &  &  & \underline{1D} &  &  & P-f \\
\rowcolor{gray!10} \cite{li_full_2022} & 15 (3) & Jilin-1, GaoFen-2 &  & 1m, 4m & Image Survey &  &  &  & \underline{2D} &  & \underline{\checkmark} & S-i \\
\cite{metzger_crop_2022} & 48 (41) & Sentinel-2 &  & 10m & ZeuriCrops$\dagger$ &  &  &  &  & L \underline{G} & T & ST-s \\
\rowcolor{gray!10} \cite{nyborg_timematch_2022} & 11 (10) & Sentinel-2 & \checkmark & 10m & Pub. govt. &  &  &  &  &  & \underline{P} & ST-o \newline (T-s) \\
\cite{paul_generating_2022} & 6 (5) & Sentinel-1 &  & 10m & Field Survey & \checkmark & \checkmark &  & \underline{2D} &  &  & ?-f \newline ST-i \\
\rowcolor{gray!10} \cite{pavlovic_monitoring_2022} & 10 (2) & Sentinel-2 &  & 10m & Image Survey &  &  &  & \underline{2D} &  &  & S-i \\
\cite{rauf_new_2022} & 3 (2) & Sentinel-2 & \checkmark & 10m & Field Survey, Model-based &  &  &  & \underline{2D} &  &  & T-i \\
\rowcolor{gray!10} \cite{rawat_comparative_2022} & 2 (1) & Sentinel-2 & \checkmark & 10m & Field Survey &  &  &  & \underline{1D} &  &  & T-s \\
\cite{saralioglu_semantic_2022} & 7 (2) & [Commercial] &  & 0.5m & Image Survey & RF & \checkmark &  & \underline{2D} \underline{3D} &  &  & P-f \newline S-i \newline S-c \\
\rowcolor{gray!10} \cite{sharma_countrywide_2022} & 101 (4) & Sentinel-2 & \checkmark & 10m & Pub. govt. &  &  &  & \underline{1D} &  &  & T-s \\
\cite{sykas_sentinel-2_2022} & 11 (11) & Sentinel-2 & \checkmark & 10m & Sen4AgriNet$\dagger$ &  &  &  & 1D 2D & 2D L & T & ST-c \newline ST-o \newline (T-s) \\
\rowcolor{gray!10} \cite{tang_channel_2022} & 13 (8) & Sentinel-2 &  & 10m & Breizhcrops$\dagger$ & RF &  &  & \underline{1D} & L & \underline{\checkmark} & T-f \newline T-s \newline ST-c \\
\cite{teimouri_fusion_2022} & 7 (7) & Sentinel-1/2 &  & 10m & Pub. govt. &  & \checkmark & \checkmark & 2D \underline{3D} &  &  & P-f \newline S-i \newline ST-c \\
\rowcolor{gray!10} \cite{wang_cctnet_2022} & 4 (3) & Sentinel-2 &  & 10m & Pub. govt. &  &  &  & \underline{2D} &  & \underline{T} & S-i \\
\cite{wang_deep_2022} & 10 (5) & Sentinel-2 & \checkmark & 10m & Field Survey &  &  &  & \underline{2D} &  &  & S-i \\
\rowcolor{gray!10} \cite{wang_evaluating_2022} & 5 (5) & MODIS & \checkmark & 1km & Model-based & \underline{RF} & \underline{\checkmark} &  & 1D & L &  & T-f \newline T-s \\
\cite{yang_fully_2022} & 4 (3) & Sentinel-2 &  & 10m & Model-based &  &  &  & \underline{1D} \underline{3D} &  &  & T-s \newline ST-c \\

\end{tabular}
}

\pagebreak

Studies on specific land cover tasks other than segmentation. The best performing model in each study is underlined.
{\small
\centering
\newcommand{\rot}{\rotatebox{90}}

\begin{tabular}{%
    >{\raggedleft}p{70pt}|
    >{\raggedleft}p{40pt}|
    >{\raggedright}p{50pt}|
    c|
    p{23pt}|
    >{\raggedright}p{55pt}|
    >{\raggedright\centering}p{12pt}|
    c|
    c|
    >{\raggedright\centering}p{12pt}|
    >{\raggedright\centering}p{12pt}|
    c|
    l
}

&  &  &  &  &  & \multicolumn{6}{c|}{\textbf{Model}} & \\
\textbf{Study} & \textbf{Task/N} & \textbf{Images} & \textbf{VI} & \textbf{GSD} & \textbf{Labels} & \rot{Tree} & \rot{SVM} & \rot{MLP} & \rot{CNN} & \rot{RNN} & \rot{Attn} & \rot{\textbf{Type}} \\
\hline

\rowcolor{white} \multicolumn{13}{l}{\textbf{Field Boundary Detection}}  \\
\hline

\cite{persello_delineation_2019} & - & WV-2/3 &  & 0.3m & Image Survey &  &  &  & \underline{2D} &  &  & S-i \\
\rowcolor{gray!10} \cite{waldner_deep_2019} & - & Sentinel-2 &  & 10m & Image Survey &  &  &  & \underline{2D} &  &  & S-i \\
\cite{masoud_delineation_2020} & - & Sentinel-2 &  & 10m & Pub. govt. &  &  &  & \underline{2D} &  &  & S-i \\
\rowcolor{gray!10} \cite{waldner_detect_2021} & - & Sentinel-2 &  & 10m & Image Survey &  &  &  & \underline{2D} &  & \underline{\checkmark} & S-i \\
\cite{jong_improving_2022} & - & MODIS, Landsat, Sentinel &  & 10m & Pub. govt., Image Survey &  &  &  & \underline{2D} &  &  & S-i \\
\rowcolor{gray!10} \cite{long_delineation_2022} & - & GF-1/2 &  & 4m & Image Survey &  &  &  & \underline{2D} &  &  & S-i \\
\cite{sharifi_agricultural_2022} & - & Sentinel-2 &  & 10m & Image Survey &  &  &  & \underline{2D} &  &  & S-i \\
\rowcolor{gray!10} \cite{mei_using_2022} & - & WV-3 &  & 0.5m & Image Survey &  &  &  & \underline{2D} &  &  & S-i \\

\rowcolor{white} \multicolumn{13}{l}{\textbf{Dam Detection}} \\
\hline
\cite{carvajal_estimating_2016} & - & Quickbird &  & 1m & ??? &  &  & \underline{\checkmark} &  &  &  & P-f \\
\rowcolor{gray!10} \cite{malerba_continental-scale_2021} & - & [Maps] &  & - & Image Survey &  &  &  & \underline{2D} &  &  & S-i \\
\cite{ma_identifying_2022} & - & GF1, GF2 &  & 15m, 4m & Image Survey &  &  &  & \underline{2D} &  &  & S-i \\
\rowcolor{gray!10} \cite{malerba_australian_2022} & - & [Maps] &  & - & Image Survey &  &  &  & \underline{2D} &  &  & S-i \\

\rowcolor{white} \multicolumn{13}{l}{\textbf{Center Pivot Irrigation Detection}} \\
\hline
\cite{zhang_automatic_2018} & Classify & Landsat-8 &  & 30m & Image Survey &  &  &  & \underline{2D} &  &  & S-i \\
\rowcolor{gray!10} \cite{de_albuquerque_deep_2020} & Segment & Landsat-8 &  & 30m & Pub. govt. &  &  &  & \underline{2D} &  &  & S-i \\
\cite{saraiva_automatic_2020} & Segment & PlanetScope &  & 3m & Image Survey &  &  &  & \underline{2D} &  &  & S-i \\
\rowcolor{gray!10} \cite{li_machine_2022} & Classify & Landsat &  & 30m & Image Survey &  &  &  & \underline{2D} &  &  & S-i \\

\rowcolor{white} \multicolumn{13}{l}{\textbf{Irrigation Detection}} \\
\hline
\cite{colligan_deep_2022} & Segment & Landsat &  & 30m & Field/Image Survey &  &  &  & \underline{2D} &  &  & S-i \\

\rowcolor{white} \multicolumn{13}{l}{\textbf{Greenhouse Detection}} \\
\hline
\cite{li_agricultural_2020} & Detection & GF1, GF2 &  & 2m, 1m & Image Survey &  &  &  & \underline{2D} &  &  & S-i \\
\rowcolor{gray!10} \cite{zhang_high-resolution_2021} & Segment Boundary & GF2 &  & 0.8m & Image Survey &  &  &  & \underline{2D} &  &  & S-i \\

\rowcolor{white} \multicolumn{13}{l}{\textbf{Hedgerow Detection}} \\
\hline
\cite{ahlswede_hedgerow_2021} & Segment & IKONOS &  & 1m & Field/Image Survey &  &  &  & \underline{2D} &  &  & S-i \\

\rowcolor{white} \multicolumn{13}{l}{\textbf{Tree Crown Delineation}} \\
\hline
\cite{gomez_use_2010} & Segment & QuickBird &  & 0.6m & ??? &  &  & \underline{\checkmark} &  &  &  & P-f \\
\rowcolor{gray!10} \cite{li_large-scale_2019} & Classify & QuickBird &  & 0.6m & Image Survey &  &  &  & \underline{2D} &  &  & S-i \\
\cite{ferreira_accurate_2021} & Segment & WV-3 &  & 0.3m & Field/Image Survey &  &  &  & \underline{2D} &  &  & S-i \\
\rowcolor{gray!10} \cite{lin_toward_2021} & Classify & WV-3, PlanetScope &  & 0.3m, 3m & Image Survey &  &  &  & \underline{2D} & L &  & S-i \\
\cite{abozeid_large-scale_2022} & Segment & Satellites.pro &  & ? & Image Survey &  &  &  &  &  & \underline{T} & S-i \\
\rowcolor{gray!10}\cite{lu_mapping_2022} & Segment & Sentinel & \checkmark & 10m & Field Survey &  &  & \underline{\checkmark} &  &  &  & P-f \\

\end{tabular}
}

\pagebreak

\subsection{All Soil Monitoring studies}

Studies on soil health; mostly soil moisture and soil salinity studies. VI stands for vegetative indices, and indicates whether VIs were used (possibly in combination with other features). GSD = Ground Spatial Distance/resolution. The best performing model type in each study is underlined. In the Tree column: RF = Random Forest, and a tick means any other kind of tree. In the CNN column: 2D = 2DCNN.
{\small
\centering
\newcommand{\rot}{\rotatebox{90}}

\begin{tabular}{>{\raggedleft}p{70pt}|>{\raggedright}p{40pt}|>{\raggedright}p{60pt}|c|p{23pt}|>{\raggedright}p{80pt}|>{\raggedright\centering}p{10pt}|c|c|c|c|p{20pt}}
&  &  &  &  &  & \multicolumn{5}{c|}{\textbf{Model}} & \\
\textbf{Study} & \textbf{Task} & \textbf{Images} & \textbf{VI} & \textbf{GSD} & \textbf{Labels} & \rot{Tree} & \rot{SVM} & \rot{MLP} & \rot{CNN} & \rot{RNN} & \textbf{Type} \\
\hline

 \rowcolor{white} \multicolumn{11}{l}{\textbf{Soil Moisture}} \\
\hline
\cite{del_frate_wheat_2004} & SM & ERS-2 &  & 30m & Model-based (train),  Field survey (eval) &  &  & \underline{\checkmark} &  &  & P-f \\
\rowcolor{gray!10} \cite{santi_comparison_2013} & SM & ASAR &  & 30m & Model-based (train),  Field survey (eval) &  &  & \underline{\checkmark} &  &  & P-f \\
\cite{baghdadi_coupling_2016} & SM & RADARSAT-2, Landsat-7/8 &  & 12m, 30m & Field Survey &  &  & \underline{\checkmark} &  &  & S-o \newline (P-f) \\
\rowcolor{gray!10} \cite{kolassa_estimating_2018} & SM & [Many] &  & 3km, 16km, 36km & Model-based (train),  Field survey (eval) &  &  & \underline{\checkmark} &  &  & P-f \\
\cite{van_der_schalie_effect_2018} & Agreement & AMSR-E, MIRAS, ASCAT &  & 25km, 35km, 25km & Self-labelled &  &  & \underline{\checkmark} &  &  & P-f \\
\rowcolor{gray!10} \cite{eroglu_high_2019} & SM & CYGNSS & \checkmark & 9km & Government &  &  & \underline{\checkmark} &  &  & P-f \\
\cite{kumar_comprehensive_2019} & SM & Sentinel-1 &  & 20m & Field Survey & RF & \underline{\checkmark} & \checkmark &  &  & P-f \\
\rowcolor{gray!10} \cite{zhang_estimation_2020} & SM & Landsat, Terra-SAR & \checkmark & 30m, 3m & Model-based (train),  Field survey (eval) &  &  & \underline{\checkmark} &  &  & P-f \\
\cite{rabiei_method_2021} & SM & Sentinel-1, Sentinel-2 &  & 10m & Field Survey &  &  & \checkmark & \underline{2D} &  & P-f \newline S-i \\
\rowcolor{gray!10} \cite{senanayake_estimating_2021} & SM & MODIS & \checkmark & 1km & Field Survey & \underline{\checkmark} &  & \checkmark &  &  & P-f \\
\cite{ghasemloo_estimating_2022} & SM & Landsat, Sentinel-1 & \checkmark & 30m, 10m & Field Survey &  &  & \underline{\checkmark} &  &  & P-f \\
\rowcolor{gray!10} \cite{tripathi_deep_2022} & SM, salinity, etc. & Sentinel-1/2 & \checkmark & 10m & Field Survey & RF \checkmark & \checkmark & \checkmark &  &  & P-f \\
\cite{xu_downscaling_2022} & SM & SMAP, MODIS & \checkmark & 36km, 1km & Model-based (train), Government (eval) &  &  & \underline{\checkmark} &  &  & P-f \\
\rowcolor{gray!10} \cite{zeynoddin_structural-optimized_2022} & SM & SMAP &  & 36km & Self-labelled &  &  &  &  & \underline{L} & T-s \\

\rowcolor{white} \multicolumn{11}{l}{\textbf{Soil Nutrients/Heavy Metals}} \\
\hline
\cite{gautam_residual_2011} & Nitrate & Landsat &  & 30m & Field Survey &  &  & \underline{\checkmark} &  &  & S-o \newline (P-f) \\
\rowcolor{gray!10} \cite{song_predicting_2018} & Nitrogen & HJ-1 &  & 100m & Field Survey & RF & \checkmark & \underline{\checkmark} &  &  & ?-f \\
\cite{muradyan_estimating_2020} & Mo, Cu, Ni, Cd & SPOT-7 &  & 1.5m & Field Survey &  &  & \underline{\checkmark} &  &  & S-o\newline (P-f) \\
\rowcolor{gray!10} \cite{zhang_temporal_2020} & Potassium & Landsat &  & 30m & Field Survey &  &  & \underline{\checkmark} &  &  & P-f \\
\cite{peng_new_2022} & Fertility & Sentinel-2 & \checkmark & 10m & Field Survey & \checkmark &  & \underline{\checkmark} &  &  & P-f \\
\rowcolor{gray!10} \cite{wang_exploring_2022} & Pb, Cd & Sentinel-2 & \checkmark & 10m & Field Survey & \underline{RF} &  & \checkmark &  &  & P-f \\
\cite{zhang_retrieving_2022} & Zn, Ni, Cu & GaoFen-5 &  & 30m & Field Survey & \underline{RF} & \checkmark & \checkmark &  &  & P-f \\

\rowcolor{white} \multicolumn{11}{l}{\textbf{Soil Salinity}} \\
\hline
\cite{wang_estimation_2018} & Regress & Landsat, HJ-1 & \checkmark & 30m, 100m & Field Survey &  &  & \underline{\checkmark} &  &  & P-f \\
\rowcolor{gray!10} \cite{qi_soil_2020} & Regress & Sentinel-2 & \checkmark & 10m & Field Survey / Model-based & \underline{RF} & \checkmark & \checkmark &  &  & P-f \\
\cite{habibi_quantitative_2021} & Regress & Landsat &  & 30m & Field Survey &  &  & \underline{\checkmark} &  &  & P-f \\
\rowcolor{gray!10} \cite{akca_semantic_2022} & Classify & RapidEye & \checkmark & 6.5m & Field Survey &  & \checkmark &  & \underline{2D} &  & S-i \\

\end{tabular}
}

\pagebreak

\subsection{All Plant Physiology studies}

Studies on plant physiology; mostly canopy cover/LAI estimates. VI stands for vegetative indices, and indicates whether VIs were used (possibly in combination with other features). GSD = Ground Spatial Distance/resolution. The best performing model type in each study is underlined. LAI = Leaf Area Index, LCC = Leaf Chlorophyl Content, LWC = Leaf Water Content, GPP = Gross Primary Production, SIF = Solar-Induced Fluorescence. In the RNN column: L = LSTM, G = GRU, and 2D = 2D ConvRNN (ConvGRU or ConvLSTM).
{\small
\centering
\newcommand{\rot}{\rotatebox{90}}

\begin{tabular}{>{\raggedleft}p{60pt}|>{\raggedright}p{55pt}|>{\raggedright}p{50pt}|c|p{20pt}|>{\raggedright}p{80pt}|c|c|c|c|>{\raggedright\centering\arraybackslash}p{12pt}|p{20pt}}

&  &  &  &  &  & \multicolumn{5}{c|}{\textbf{Model}} & \\
\textbf{Study} & \textbf{Task} & \textbf{Images} & \textbf{VI} & \textbf{GSD} & \textbf{Labels} & \rot{Tree} & \rot{SVM} & \rot{MLP} & \rot{CNN} & \rot{RNN} & \textbf{Type} \\
\hline

\rowcolor{white} \multicolumn{12}{l}{\textbf{Canopy Cover/LAI}} \\
\hline
\cite{del_frate_wheat_2004} & Regress LAI & ERS-2 &  & 30m & Model-based (train),  Field survey (eval) &  &  & \underline{\checkmark} &  &  & P-f \\
\rowcolor{gray!10} \cite{gascon_using_2007} & Regress LAI & POLDER &  & 60m & Model-based (train),  Field survey (eval) &  &  & \underline{\checkmark} &  &  & P-f \\
\cite{bsaibes_albedo_2009} & Regress LAI & FORMOSAT-2 &  & 8m & Field Survey &  &  & \underline{\checkmark} &  &  & S-o\newline (P-f) \\
\rowcolor{gray!10} \cite{richter_experimental_2009} & Regress LAI & CASI &  & 10m & Model-based (train),  Field survey (eval) &  &  & \underline{\checkmark} &  &  & P-f \\
\cite{verger_optimal_2011} & Regress LAI / Canopy Cover & PROBA-1 &  & 34m & Model-based (train),  Field survey (eval) &  &  & \underline{\checkmark} &  &  & P-f \\
\rowcolor{gray!10} \cite{bocco_estimating_2012} & Regress Canopy Cover & MODIS, Landsat & \checkmark & 500m, 30m & Field Survey &  &  & \underline{\checkmark} &  &  & S-o\newline (P-f) \\
\cite{wu_comparison_2015} & Regress LAI & HJ-1 & \checkmark & 30m & Field Survey &  &  & \underline{\checkmark} &  &  & P-f \\
\rowcolor{gray!10} \cite{baghdadi_coupling_2016} & Regress LAI & RADARSAT-2, Landsat-7/8 &  & 12m, 30m & Field Survey &  &  & \underline{\checkmark} &  &  & S-o\newline (P-f) \\
\cite{kira_toward_2017} & Regress LAI & MODIS, Landsat, MERIS &  & 500m, 30m, 250m & Field Survey &  & \underline{\checkmark} & \checkmark &  &  & S-o\newline (P-f) \\
\rowcolor{gray!10} \cite{delloye_retrieval_2018} & Regress LAI & Sentinel-2 &  & 10m & Model-based (train),  Field survey (eval) &  &  & \underline{\checkmark} &  &  & P-f \\
\cite{sun_leaf_2021} & Regress LAI & MODIS &  & 500m & Model-based (train),  Field survey (eval) &  &  & \underline{\checkmark} &  &  & P-f \\
\rowcolor{gray!10} \cite{tomicek_prototyping_2021} & Regress LAI, LCC, LWC & Sentinel-2 &  & 10m & Model-based (train),  Field survey (eval) &  &  & \underline{\checkmark} &  &  & P-f \\
\cite{elmetwalli_assessing_2022} & Regress LAI, height, AGB, SPAD & QuickBird & \checkmark & 0.6m & Field Survey & RF &  & \underline{\checkmark} &  &  & P-f \\
\rowcolor{gray!10} \cite{igder_multivariate_2022} & Regress LAI & Sentinel-2 &  & 10m & Model-based &  &  & \underline{\checkmark} &  &  & P-f \\

\rowcolor{white} \multicolumn{12}{l}{\textbf{Other}} \\
\hline
\cite{wagle_parameterizing_2016} & Regress Light / Water Use Efficiency & MODIS & \checkmark & 500m & Government &  &  & \underline{\checkmark} &  &  & T-f \\
\rowcolor{gray!10} \cite{wolanin_estimating_2019} & Regress GPP & Sentinel-2 & \checkmark & 10m & Model-based (train),  Field survey (eval) &  &  & \underline{\checkmark} &  &  & P-f \\
\cite{kira_extraction_2020} & Regress SIF & MODIS &  & 500m & Government &  &  & \underline{\checkmark} &  &  & P-f \\
\rowcolor{gray!10} \cite{thorp_deep_2021} & Classify Growth Stage & Sentinel-1, Sentinel-2 & \checkmark & 10m & Government & RF & \checkmark & \checkmark & 2D & \underline{L} G 2D  & P-f\newline S-i\newline T-s\newline ST-c \\
\cite{zhao_spatial-aware_2022} & Classify Growth Stage & Sentinel-1/2 & \checkmark & 10m & Field survey, Model-based &  &  &  & 2D &  & S-i \\

\end{tabular}
}

\pagebreak

\subsection{All Crop Damage studies}

Studies on crop damage. VI stands for vegetative indices, and indicates whether VIs were used (possibly in combination with
other features). GSD = Ground Spatial Distance/resolution. In the Tree column: RF = Random Forest, and a tick means any
other kind of tree. In the RNN column: G = GRU. The best performing model type in each study is underlined.
{\small
\centering
\newcommand{\rot}{\rotatebox{90}}

\begin{tabular}{>{\raggedleft}p{70pt}|>{\raggedright}p{75pt}|p{50pt}|c|p{20pt}|p{50pt}|c|c|c|c|c|p{20pt}}

& &  &  &  &  & \multicolumn{5}{c|}{\textbf{Model}} & \\
\textbf{Study} & \textbf{Task} & \textbf{Images} & \textbf{VI} & \textbf{GSD} & \textbf{Labels} & \rot{Tree} & \rot{SVM} & \rot{MLP} & \rot{CNN} & \rot{RNN} & \textbf{Type} \\
\hline

\rowcolor{white} \multicolumn{12}{l}{\textbf{Disease}} \\
\hline
\cite{yuan_damage_2014} & Segment powdery mildew & Spot-6 & \checkmark & 1.5m & Field Survey &  &  & \underline{\checkmark} &  &  & S-o\newline (P-f) \\
\rowcolor{gray!10} \cite{ma_integrating_2019} & Segment powdery mildew and aphids & Landsat & \checkmark & 30m & Field Survey &  & \checkmark & \underline{\checkmark} &  &  & P-f \\
\cite{bi_gated_2020} & Segment SDS in soybeans & PlanetScope &  & 3m & Field Survey & \checkmark &  & \checkmark &  & \underline{G} & T-f\newline T-s \\
\rowcolor{gray!10} \cite{pignatti_sino-eu_2021} & Segment P. syringae and Yellow Rust & Sentinel-2, RapidEye-1, PRISMA & \checkmark & 10m, 6.5m, 30m & Field Survey &  &  & \underline{\checkmark} &  &  & S-o\newline (P-f) \\
\cite{ruan_prediction_2021} & Segment wheat stripe rust & Sentinel-2 & \checkmark & 10m & Field Survey &  & \underline{\checkmark} & \checkmark &  &  & P-f \\
\rowcolor{gray!10} \cite{guo_recognition_2022} & Segment Yellow Leaf Disease & PlanetScope & \checkmark & 3.1m & Field Survey & \underline{RF} &  & \checkmark &  &  & P-f \\

\rowcolor{white} \multicolumn{12}{l}{\textbf{Generic/Physical Damage}} \\
\hline
\rowcolor{gray!10} \cite{rodriguez_robust_2021} & Regress coconut tree density & Sentinel-2 &  & 10m & Image Survey &  &  &  & \underline{2D} &  & S-i \\
\cite{virnodkar_denseresunet_2021} & Segment water stress & Sentinel-2 &  & 10m & Field Survey &  &  &  & \underline{2D} &  & S-i \\
\rowcolor{gray!10} \cite{boroughani_assessment_2022} & Segment dusty day & MODIS &  & 250m & Field/Image Survey & \underline{\checkmark} &  & \checkmark &  &  & P-f \\

\end{tabular}
}

\pagebreak

\subsection{All Yield Estimation studies}

The majority of the yield estimation studies were county-level. 

VI stands for vegetative indices, and indicates whether VIs were used (possibly in combination with other features). GSD = Ground Spatial Distance/resolution. ``Clim.?'' indicates whether additional climate data was used in the prediction. The best performing model type in each study is underlined. Where two model types are underlined, it means the model had components of both types. In the Tree column: RF = Random Forest, and a tick means any other kind of tree. In the CNN column: nD = nDCNN, including non-spatial CNNs. In the RNN column: L = LSTM
{
\small
\centering

\newcommand{\rot}{\rotatebox{90}}

\begin{tabular}{%
    >{\raggedleft}p{70pt}|
    >{\raggedright}p{40pt}|
    >{\raggedright}p{40pt}|
    >{\raggedright\centering}p{12pt}|
    p{20pt}|
    >{\raggedright\centering}p{12pt}|
    >{\raggedright}p{55pt}|
    >{\raggedright\centering}p{12pt}|
    >{\raggedright\centering}p{12pt}|
    >{\raggedright\centering}p{12pt}|
    >{\raggedright\centering}p{12pt}|
    >{\raggedright\centering\arraybackslash}p{12pt}|
    p{23pt}
}
 &  &  &  &  &  &  & \multicolumn{5}{c|}{\textbf{Model}} & \\
\textbf{Study} & \textbf{Crop} & \textbf{Images} & \textbf{VI} & \textbf{GSD} & \rot{\textbf{Clim.?}} & \textbf{Labels} & \rot{Tree} & \rot{SVM} & \rot{MLP} & \rot{CNN} & \rot{RNN} & \textbf{Type} \\
\hline

\rowcolor{white} \multicolumn{13}{l}{\textbf{County-level yield}} \\
\hline
\cite{li_estimating_2007} & Maize, Soy & MODIS, AVHRR & \checkmark & 1km &  & Pub. govt. &  &  & \underline{\checkmark} &  &  & ST-o \newline (T-f) \\
\rowcolor{gray!10} \cite{cai_integrating_2019} & Wheat & MODIS & \checkmark & 5.5km & \checkmark & Pub. govt. & RF & \underline{\checkmark} & \checkmark &  &  & ST-o \newline (T-s) \\
\cite{feng_machine_2019} & Wheat & MODIS & \checkmark & 1km & \checkmark & Pub. govt. & \underline{RF} & \checkmark & \checkmark &  &  & ST-o \newline (P-f) \\
\rowcolor{gray!10} \cite{kang_comparative_2020} & Maize & MODIS, [Many] & \checkmark & 500m & \checkmark & Pub. govt. & RF \checkmark & \checkmark &  & 2D & \underline{L} & ST-o \newline (T-f \newline T-s \newline T-i) \\
\cite{potopova_statistical_2020} & Maize, Sunflower, Grapes & MODIS & \checkmark & 5km & \checkmark & Pub. govt. &  &  & \underline{\checkmark} &  &  & ST-o \newline (T-f) \\
\rowcolor{gray!10} \cite{schwalbert_satellite-based_2020} & Soy & MODIS & \checkmark & 250m & \checkmark & Pub. govt. & RF &  &  &  & \underline{L} & ST-o \newline (T-f \newline T-s) \\
\cite{wang_combining_2020} & Wheat & MODIS & \checkmark & 500m & \checkmark & Pub. govt. & RF \underline{\checkmark} & \checkmark & \checkmark &  &  & ST-o \newline (T-f) \\
\rowcolor{gray!10} \cite{wolanin_estimating_2020} & Wheat & MODIS & \checkmark & 500m & \checkmark & Pub. govt. & RF &  &  & \underline{1D} &  & ST-o \newline (T-f \newline T-s) \\
\cite{zhang_combining_2020} & Maize & MODIS & \checkmark & 1km & \checkmark & Pub. govt. & RF \underline{\checkmark} &  &  &  & L & ST-o \newline (T-f \newline T-s) \\
\rowcolor{gray!10} \cite{cao_integrating_2021} & Rice & MODIS & \checkmark & 1km & \checkmark & Pub. govt. & RF &  &  &  & \underline{L} & ST-o \newline (T-f \newline T-s) \\
\cite{cao_wheat_2021} & Wheat & MODIS & \checkmark & 1km & \checkmark & Pub. govt. & \underline{RF} &  & \checkmark & 1D & L & ST-o \newline (T-f \newline T-s) \\
\rowcolor{gray!10} \cite{feng_geographically_2021} & Wheat & MODIS & \checkmark & 500m & \checkmark & Pub. govt. &  &  & \underline{\checkmark} &  &  & ST-o \newline (T-f) \\
\cite{ju_optimal_2021} & Maize, Soy, Rice & MODIS & \checkmark & 250m, 1km & \checkmark & Pub. govt. & RF & \underline{\checkmark} & \checkmark & 2D & L & ST-o \newline (T-f \newline T-s) \\
\rowcolor{gray!10} \cite{khaki_simultaneous_2021} & Maize, Soy & MODIS &  & 1km &  & Pub. govt. & RF &  & \checkmark & \underline{2D} 3D &  & ST-o \newline (T-f \newline T-i \newline T-c) \\
\cite{ma_corn_2021} & Maize & MODIS & \checkmark & 500m & \checkmark & Pub. govt. &  &  & \underline{\checkmark} &  &  & ST-o \newline (T-f) \\
\rowcolor{gray!10} \cite{tian_deep_2021} & Wheat & MODIS & \checkmark & 1km & \checkmark & Pub. govt., Field Survey &  &  &  &  & \underline{L} & ST-o \newline (T-s) \\
\cite{qiao_exploiting_2021} & Wheat & MODIS &  & 500m &  & Pub. govt. &  &  &  & \underline{3D} &  & ST-o \newline (T-c) \\
\rowcolor{gray!10} \cite{xie_integration_2021} & Wheat & MODIS & \checkmark & 250m & \checkmark & Model-based, Pub. govt. & RF &  &  & 1D & \underline{L} & ST-o \newline (T-f \newline T-s) \\
\cite{ji_prediction_2022} & Maize & MODIS & \checkmark & 1km & \checkmark & Pub. govt. & RF &  &  &  & \underline{L} & ST-o \newline (T-f \newline T-s) \\
\rowcolor{gray!10} \cite{liu_exploring_2022} & Wheat & MODIS & \checkmark & 1km & \checkmark & Pub. govt. & RF \checkmark & \underline{\checkmark} &  &  & L & ST-o \newline (T-f \newline T-s) \\
\cite{luo_accurately_2022} & Wheat & AVHRR, MODIS & \checkmark & 1km & \checkmark & Pub. govt. & RF \checkmark &  &  &  & \underline{L} & ST-o \newline (T-f \newline T-s) \\
\rowcolor{gray!10} \cite{watson-hernandez_oil_2022} & Palm Oil & Landsat-5/6/7/8 & \checkmark & 15m, 30m & \checkmark & Business & RF \checkmark &  & \checkmark &  &  & ST-o \newline (T-f) \\
\cite{xie_combining_2022} & Wheat & Sentinel-2 & \checkmark & 10m & \checkmark & Model-based, Pub. govt. &  &  &  &  & \underline{L} & ST-o \newline (T-s) \\

\end{tabular}
}

\pagebreak

There were a few studies attempting to predict yield at different scales, too, but not many.

{
\small
\centering

\newcommand{\rot}{\rotatebox{90}}

\begin{tabular}{%
    >{\raggedleft}p{70pt}|
    >{\raggedright}p{40pt}|
    >{\raggedright}p{50pt}|
    >{\raggedright\centering}p{12pt}|
    p{20pt}|
    >{\raggedright\centering}p{12pt}|
    >{\raggedright}p{50pt}|
    >{\raggedright\centering}p{12pt}|
    >{\raggedright\centering}p{12pt}|
    >{\raggedright\centering}p{12pt}|
    >{\raggedright\centering}p{12pt}|
    >{\raggedright\centering\arraybackslash}p{12pt}|
    p{20pt}
}
 &  &  &  &  &  &  & \multicolumn{5}{c|}{\textbf{Model}} & \\
\textbf{Study} & \textbf{Crop} & \textbf{Images} & \textbf{VI} & \textbf{GSD} & \rot{\textbf{Clim.?}} & \textbf{Labels} & \rot{Tree} & \rot{SVM} & \rot{MLP} & \rot{CNN} & \rot{RNN} & \textbf{Type} \\
\hline

\rowcolor{white} \multicolumn{13}{l}{\textbf{Field-level yield}} \\
\hline
\cite{khan_artificial_2020} & Mentha & Landsat & \checkmark & 30m &  & Field Survey &  &  & \underline{\checkmark} &  &  & S-o \newline (P-f) \\
\rowcolor{gray!10} \cite{arab_prediction_2021} & Grapes & Landsat & \checkmark & 30m &  & Field Survey &  &  & \underline{\checkmark} &  &  & S-o \newline (P-f) \\
\cite{evans_long-term_2021} & Wheat & Landsat & \checkmark & 30m & \checkmark & Field Survey & RF & \checkmark & \underline{\checkmark} &  &  & ST-o \newline (P-f) \\
\rowcolor{gray!10} \cite{gbodjo_benchmarking_2021} & Millet & Sentinel, Planetscope & \checkmark & 10m, 3.1m &  & Field Survey & RF &  & \underline{\checkmark} & 2D & L & ST-o \newline (T-f \newline T-s \newline ?-i) \\
\cite{zhang_integrating_2021} & Maize & Landsat & \checkmark & 30m & \checkmark & Priv. govt. & \checkmark &  &  &  & \underline{L} & ST-o \newline (T-f \newline T-s) \\
\rowcolor{gray!10} \cite{abebe_combined_2022} & Sugarcane & Sentinel-2, Landsat & \checkmark & 30m &  & Field Survey &  & \underline{\checkmark} & \checkmark &  &  & ST-o \newline (P-f) \\
\cite{krupavathi_field-scale_2022} & Sugarcane & Landsat & \checkmark & 30m &  & Field Survey &  &  & \underline{\checkmark} &  &  & ST-o \newline (P-f) \\

\rowcolor{white} \multicolumn{13}{l}{\textbf{Farm-level yield}} \\
\hline
\cite{engen_farm-scale_2021} & Any & Sentinel-2 & \checkmark & 10m & \checkmark & Pub. govt. &  &  & \checkmark & \underline{2D} & L \underline{G} & ST-c \\

\rowcolor{white} \multicolumn{13}{l}{\textbf{Plot-level yield}} \\
\hline
\cite{haghverdi_prediction_2018} & Cotton & Landsat & \checkmark & 30m & \checkmark & Field Survey &  &  & \underline{\checkmark} &  &  & T-f \\
\rowcolor{gray!10} \cite{sagan_field-scale_2021} & Maize, Soy & WorldView-3, PlanetScope &  & 0.3m, 3.1m &  & Field Survey & RF & \checkmark & \checkmark & 2D \underline{3D} &  & ST-o \newline (P-f) \newline ST-c \\

\rowcolor{white} \multicolumn{13}{l}{\textbf{Tree-level yield}} \\
\hline
\cite{rahman_exploring_2018} & Mango & Worldview-3 & \checkmark & 0.3m &  & Field Survey &  &  & \underline{\checkmark} &  &  & S-i \\

\rowcolor{white} \multicolumn{13}{l}{\textbf{Pixel-level yield}} \\
\hline
\cite{jeong_predicting_2022} & Rice & MODIS & \checkmark & 1km & \checkmark & Model-based &  &  &  & \underline{1D} & \underline{L} & T-s \\
\rowcolor{gray!10} \cite{tripathi_deep_2022} & Wheat & Sentinel-1/2 & \checkmark & 10m &  & Priv. govt. &  &  & \underline{\checkmark} &  &  & P-f \\

\rowcolor{white} \multicolumn{13}{l}{\textbf{Grassland Biomass}} \\
\hline
\cite{li_mapping_2016} & Grass & Landsat &  & 30m &  & Field Survey &  &  & \underline{\checkmark} &  &  & P-f \\
\rowcolor{gray!10} \cite{ali_modeling_2017} & Grass & MODIS & \checkmark & 500m &  & Field Survey &  &  & \underline{\checkmark} &  &  & P-f \\
\cite{barnetson_climate-resilient_2021} & Grass & PlanetScope & \checkmark & 3m &  & Field Survey &  &  & \underline{\checkmark} &  &  & T-f \\
\rowcolor{gray!10} \cite{nickmilder_development_2021} & Grass & Sentinel-1/2 &  & 10m & \checkmark & Field Survey & RF & \checkmark & \checkmark &  &  & T-f \\

\rowcolor{white} \multicolumn{13}{l}{\textbf{County-level cumulative NDVI}} \\
\hline
\cite{zambrano_prediction_2018} & All & MODIS & \checkmark & 500m, 18km & \checkmark & Images &  &  & \underline{\checkmark} &  &  & P-f \\

\end{tabular}
}

\pagebreak

\bibliographystyle{plainnat}
\bibliography{references}